# Survey of Trustworthy AI: A Meta Decision of AI

Caesar Wu[a] ∗, Yuan-Fang Li[b], and Pascal Bouvry[a]

[a] *SnT & Department of Computer Science/FSTM/University of Luxembourg, 6 Avenue de la Fonte, L-4364 Esch-sur-Alzette, Luxembourg*
[b] *Faculty of Information Technology/Monash University, 20 Exhibition Walk, Clayton, Vic 3800, Australia*

ABSTRACT

When making strategic decisions, we are often confronted with overwhelming information to process. The situation can be further complicated when some pieces of evidence are contradicted each other or paradoxical. The challenge then becomes how to determine which information is useful and which ones should be eliminated. This process is known as meta-decision. Likewise, when it comes to using Artificial Intelligence (AI) systems for strategic decision-making, placing trust in the AI itself becomes a meta-decision, given that many AI systems are viewed as opaque "black boxes" that process large amounts of data. Trusting an opaque system involves deciding on the level of Trustworthy AI (TAI). We propose a new approach to address this issue by introducing a novel taxonomy or framework of TAI, which encompasses three crucial domains: articulate, authentic, and basic for different levels of trust. To underpin these domains, we create ten dimensions to measure trust: explainability/transparency, fairness/diversity, generalizability, privacy, data governance, safety/robustness, accountability, reproducibility, reliability, and sustainability. We aim to use this taxonomy to conduct a comprehensive survey and explore different TAI approaches from a strategic decision-making perspective. We adopt the Preferred Reporting Items for Systematic Review and Meta-Analysis (PRISMA) methodology to investigate different TAI schemes and tools. This method allows us to evaluate various TAI implementation models through three lenses of ethical theories: utility-based (utilitarianism), duty-based (deontology), and natural law-based. Moreover, we discuss some ethics, machine learning, and strategic decision-making paradoxes along the meta-decision process. By acknowledging these paradoxes, we can strive towards developing better TAI approaches that align with ethical standards and advance TAI. Finally, we provide insights into the future direction in the field of TAI research.

∗ Corresponding author. Tel.: +352-466-644-9508; e-mail: Caesar.wu@uni.lu

# 1. Introduction

Determining how much trust to place in a result generated by Artificial Intelligence (AI) / Machine Learning (ML) algorithms can be an immensely daunting task, especially for a strategic decision. There are nine compelling reasons: **1.)** The nature of learning logic is reversed [1], in which the machine learns rules or patterns derived from data rather than being explicitly programmed. **2.)** While David Hume argues that we have no reason to expect the future resembles the past [2], we still intend to use historical data to make predictions about our future. **3.)** Many ML models, such as deep neural networks, support vector machines, and ensemble methods, are considered to be black boxes. Due to their lack of transparency, many black box models could pose a significant risk to a firm's strategy or high-stakes decisions [72]. **4.)** The underlying principle of ML is to identify correlations rather than causation. Pearl [4] argues that the traditional statistical approach, which focuses on identifying correlations between variables, is insufficient for understanding causality. **5.)** AI systems' security and privacy protections are critical, as cyber-attacks can result in severe consequences such as data theft, financial loss, and security breaches. These issues can significantly impact people's trust in AI systems [5]. **6.)** The lack of meaningful human control over AI systems is a significant barrier to building trustworthy AI [6]. **7.)** The subtlety of ethical issues raises another significant challenge to trustworthy AI because ethical values are cultural norms and belief-bounded [7]. **8.)** Strategic decision-making means long-term consequences for an organization. These decisions are typically made based on assumptions about the future. As a result, it can be difficult to verify the consequences immediately.[8] **9.)** Navigating the complexities of strategic decision-making can be further complicated by the paradoxical outcomes arising from different AI/ML techniques [9] and strategic objectives with competing demands. [10]

To systematically solve these challenges, data quality is a good starting point because data is a pivotal point of a solution for ML. Traditionally, we have used a rule-based approach known as good old-fashioned artificial intelligence (GOFAI) systems. With GOFAI, we define a set of rules with a given dataset and let a machine work out a desirable outcome. With ML or deep neural nets, we reverse the learning logic, in which we specify what we want with a given dataset and let a machine find rules or patterns in the representation space on our behalf. Moreover, ML applications for strategic decision-making are prevalent. They can be found in many industries and domains, such as business analysis, managerial economics, organization planning, organization behaviour, financial investment, customer segments, assets management, supply chain management, product lifecycle, market analysis, and competitive analysis.[11]

When working with a dataset that is free of corruption, bias, missing values, and inconsistency, we can have relative confidence in the accuracy of learning outcomes. Unfortunately, the data we access is often messy or incomplete, prohibiting our ability to make accurate predictions or assessments. The Data Warehousing Institute (TDWI) reported that poor data quality cost U.S. businesses $600 billion annually in 2012[12]. Recently, IBM [13] indicated that the average cost of poor-quality data would reach an average of $15 million per business or more than $3.1 trillion annually in 2022. Batini and Scannapieco [14] highlighted eight common concerns regarding data quality: 1.) accuracy, 2.) completeness, 3.) redundancy, 4.) readability, 5.) accessibility and availability, 6.) consistency, 7.) usefulness, and 8.) trust. In addition, we may also encounter issues of missing attributes, implausible values, ambiguous meanings, and untraceable data. Altogether, we can categorize these issues into two large clusters: unavailable and unusable data (See figure 1). When data is unclean and biased, the results of ML algorithms become less reliable, making it challenging to trust the outcomes.

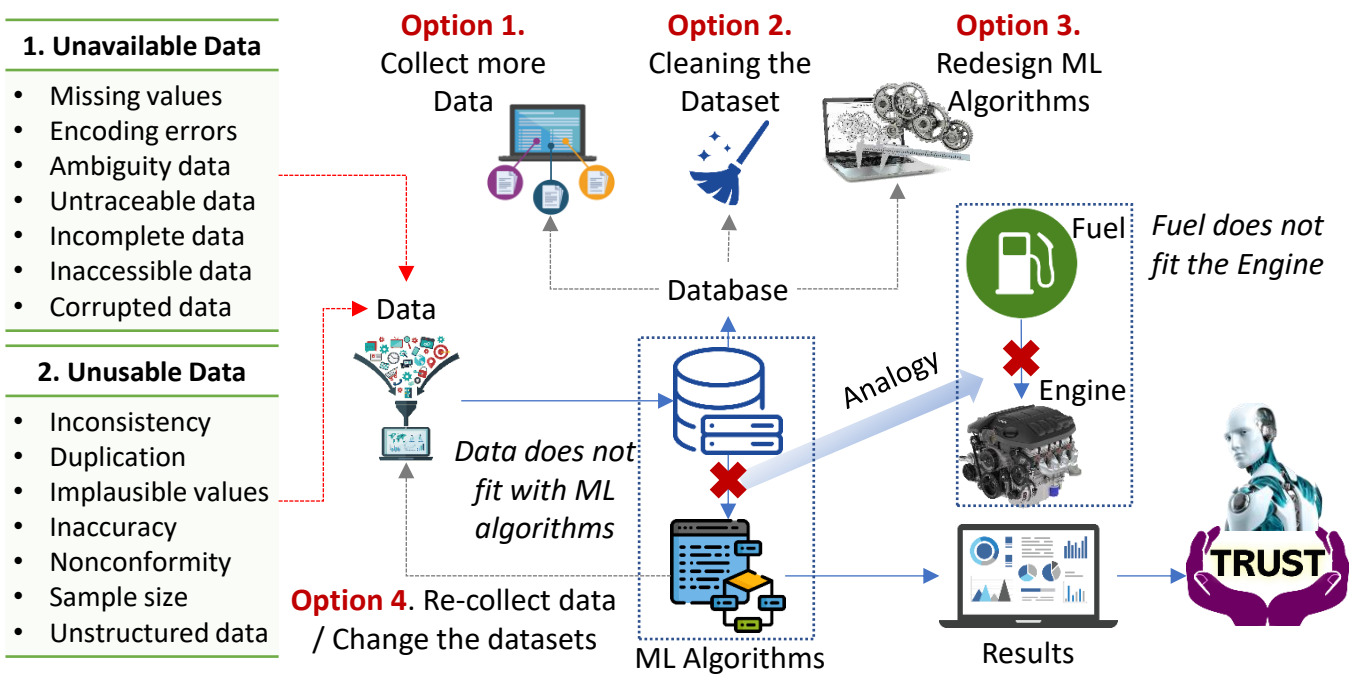

Figure 1 Data Quality and Trustworthy AI

To ensure that AI systems are trustworthy, Figure 1 illustrates four possible options in terms of data quality: 1.) Collect more data, 2.) Cleaning the dataset, 3.) Redesign ML algorithms, and 4.) Change the datasets or recollect data. Options 1 and 4 are often very expansive and sometimes impossible. Even if we make many efforts to ensure that data (fuel) fits with the learning algorithms (engine), options 2 and 3 could still end up with "garbage in and garbage out". Although data quality is one of the critical issues for TAI, the real problem sometimes could be beyond data quality because the notion of trust and trustworthiness is often emotional, relational, interactive, and dynamic[15]. Given the complexity of the TAI issue, we propose a TAI framework as a systemic or architectural solution. It can also be considered a TAI taxonomy.

The proposed TAI hierarchy consists of three levels of data processing: objectives (goal), domains (knowledge), and dimensions (measurements) (See Figure 2). The origin of this framework is derived from Marr's "levels of analysis or explanation" [21]. Marr argues that the visual system can be understood as a series of computational processes, including object recognition, feature detection, and image processing. Similarly, we can use isomorphism to formulate a framework comprising objectives, three domains and ten dimensions for TAI. (More details are in Figures 5 and 8)

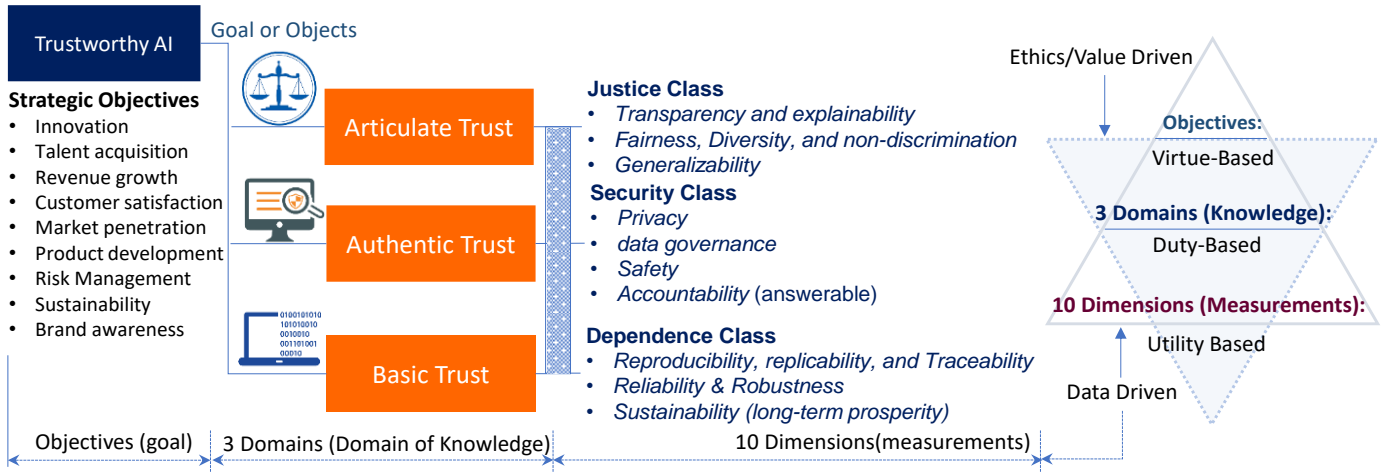

Figure 2 The Proposed TAI Hierarchical Framework

Placing the objectives at the top is justified by the fact that our moral values and ethics frequently shape our goals. The values are drawn from three main ethical theories: utilitarianism (utility-based), deontology (duty-based), and natural law (virtue-based). It creates an inverted pyramid as our guiding principles of objectives, domains and dimensions (more details in the later sections). Our TAI goal outlines the desired objectives for a particular application or a type of problem to be resolved.

The next level below is the domain upheld by domain knowledge. These domains are divided into articulate, authentic, and basic trust. They serve the purpose of different trustworthy applications or objectives. Each application (e.g., business analysis, market forecasting, self-driving cars, healthcare, online shopping, and banking services) requires particular domain knowledge. The relationship between domains and dimensions is adaptable. We can select different sets of dimensions to form a particular domain. A dimension class is just a placeholder. Objectives, domains, and dimensions generate a pyramid shape of an abstract structure to illustrate different learning processes from data. The right side of Figure 2 illustrates that this hierarchical framework is driven by both top-down (ethics/values) and bottom-up (data/facts). The framework highlights that we can either fix the data if the issue is regarding data or justify our value proposition if our value or belief does not align with the reality or facts.

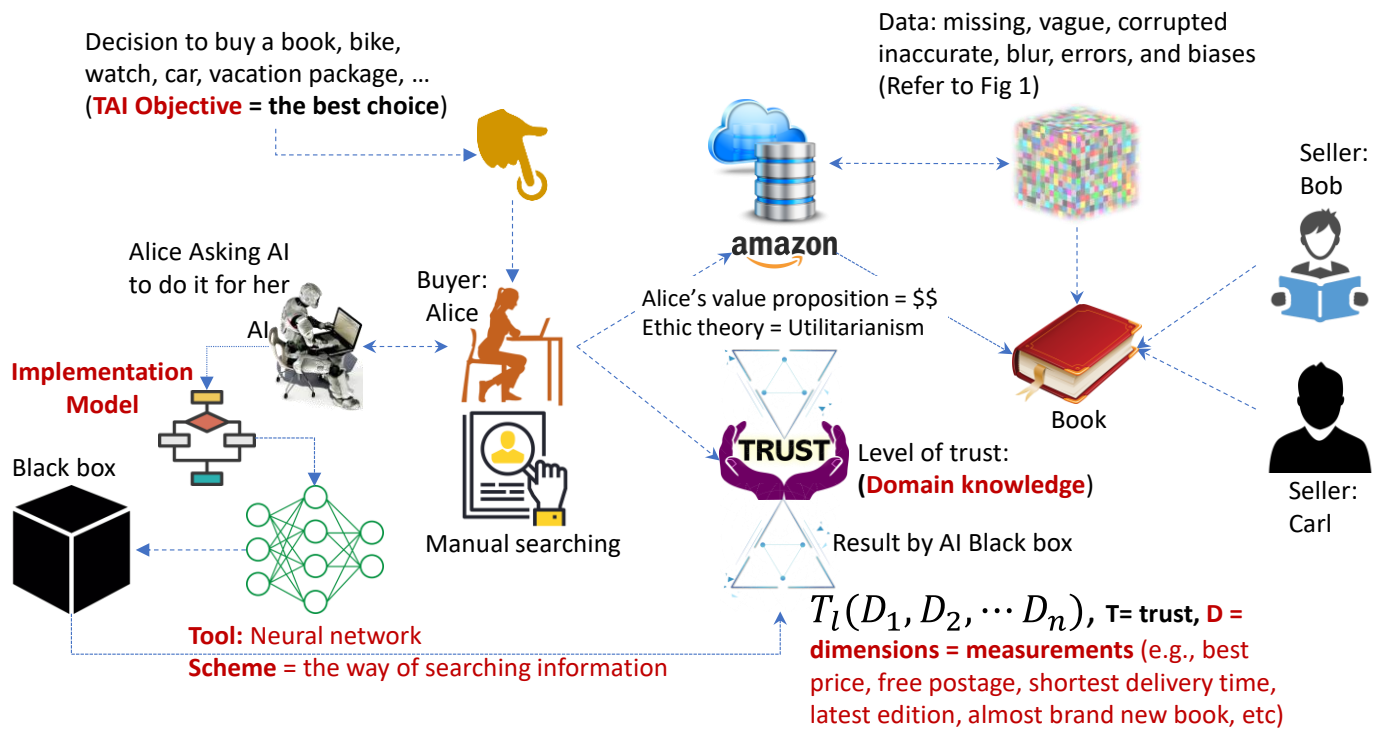

Figure 3 An Example of Explaining Taxonomy's Key Terms

To better understand these abstractive terminologies, let us look at a simple example. If Alice wants to buy a book (it can be a strategic decision, such as a hundred million dollar investment), she can search for information about the book or a seller by herself via a website or ask the AI machine for help. The TAI objective is the best choice. The domain is equivalent to domain knowledge or expertise in buying decisions. Alice might consult some domain experts with whom she has a certain level of trust, but she has her own value proposition about the book. The dimension measures the objective (e.g., the best price, the latest edition, free postage cost, shortest delivery time, etc.). We can also assume 1.) the machine uses the neural network (black box), 2.) the collecting data is messy. The question is how to trust the result generated by the machine (See Figure 3)

In order to trust the AI result, we can establish a framework or taxonomy of a "levels of analysis (or trust)" structure to investigate different TAI techniques. The taxonomy enables us to:

- Examine the pros and cons of various TAI techniques implemented by ML.
- Analyze the impact of different ML factors on the decision-making process.
- Evaluate how certain common ML paradoxes can potentially affect strategic decision-making.
- Identify and analyze possible areas for improvement based on the survey results.
- Highlight the future directions in this field of research.

*1.1. Survey Methodology*

The survey method combines Preferred Reporting Items for Systematic Reviews and Meta-Analysis (PRISMA), a list of 27 checking points classified into seven domains [16] with text-mining techniques [17][18][19] via R to screen out over 1000 literature and focus on more than about one hundred papers. Our survey method has four processing steps: 1.) The inductive process is to expand the literature pool based on TAI keywords, 2.) The text mining process is to identify the most relevant literature, 3.) The deductive process is to determine seminal papers according to the quality of the paper, 4.) The iterative process is to update and revise the latest publications during the survey process (See Figure 4).

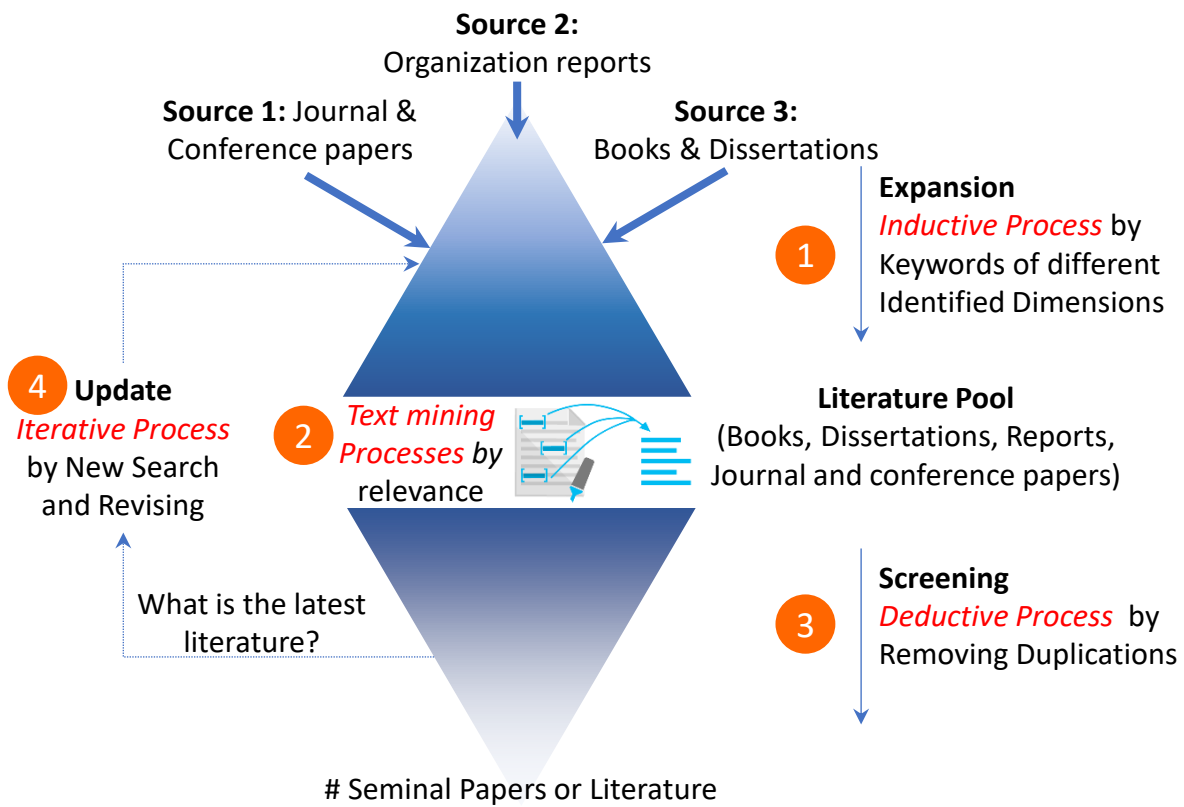

Figure 4 Survey Methodology [1]

The taxonomy method is mainly phenetic. The phenetic taxonomy compares the overall similarity between objects, and those with the highest degree of similarity are grouped together [20]. This method is based on the applications and characteristics of TAI techniques.

---

[1] Notice that we adopt the inductive process to select various types of literature from three different sources: 1.) Journal and conference papers, 2.) Organization reports and whitepapers, 3.) Books and dissertations. Generally speaking, a typical conference paper has between 8 and 15 pages, while a journal paper usually has between 12 and 35 pages. The regular size of a report is between 50 and 150 pages or less than 200 pages. The average length of a book or PhD dissertation is more than 200 pages. We use this classification to run the initial text mining process, which can identify the most relevant literature in the collected literature pool.

*1.2. Primary Contributions*

With the designated taxonomy and survey method, we choose literature from at least three sources (including conference and journal papers, books, technical reports, whitepapers, and PhD dissertations) and summarize selected papers regarding their research method, contributions, and future improvement for each TAI technique. Consequently, we made the following contributions:

- We establish a compelling taxonomy encompassing objectives, three domains, ten dimensions, 33 implementation models, and 106 schemes. The dimensions serve as primary metrics for the TAI techniques evaluation. The implementation models are the solutions to various problems, while schemes are the tools of the ML toolbox.

- We offer a comprehensive survey of the states of the arts for TAI techniques regarding their contributions to TAI.

- The survey method combines PRISMA with the text-mining approach to select relevant literature.

- The framework can be adapted and adopted as a TAI computational solution.

- This study identifies various Ethics, ML, and strategic decision paradoxes and dilemmas due to the complexity of TAI.

*1.3. Scope of the Research*

The rest of the survey is organized as follows. **Section 2** begins with clarifying essential concepts regarding trust and trustworthiness and their relationship. These concepts divide into three layers: objectives, domains, and dimensions based on David Marr's theory of the vision information process.[22] **Section 3** presents the details of the novel taxonomy comprising 3 domains, 10 dimensions, 33 implementation models, and 106 schemes. The core of the taxonomy is the trust domain, which is divided into three levels. **Section 4** provides a comprehensive survey regarding TAI. **Section 5** briefly discusses human decisions on TAI derived from three ethical theories: 1.) utility-based, 2.) duty-based, and 3.) natural law-based theory, and then identifies some ethical AI, ML, TAI, and strategic decision paradoxes that may impact strategic decision-making. **Section 6** summarizes and discusses TAI at different levels and provides conclusions and future research direction.

**2. Related Terminologies of Trust**

The decision on TAI is very challenging, especially for an application of high-stake decision, because it involves many aspects of our subjective views, such as understanding AI, human beliefs, experiences, ethical values, emotions, justice, Fairness, equality, duty, right and wrong, and good and bad. [23] Many items are tough to be quantified. Deciding TAI is ultimately a question of deciding how to decide. As the famous legal maxim goes, 'hard cases make bad law,' so we start by delving into the simple TAI concepts before deciding how to decide.

*2.1. Objectives Terminologies Trust and Trustworthiness*

The TAI framework is designed to capture specific requirements and represent particular domain knowledge for some objectives. It is built on some essential concepts, such as trust, trustworthiness, AI, decision, ethics, value, and policy. Flores and Solomon[15] argued that 'trust' itself is a decision. It determines relationship bonds towards someone or something. Considering the societal perspective, trust is not only relational but also carries significant emotional weight. Like all emotional feelings, trust is dynamic. It never means a single instance of trust. Consequently, trust can be either created or restored or destroyed or broken. Similarly, trustworthiness carries the same meaning.

We often use these two terms interchangeably. Sometimes, these two terms are confusing because they are closely related concepts. Lexically, trust means belief in reliability. Flores and Solomon argued [15] that trust is often the result of something being perceived as trustworthy. These two terms form an obvious complementary pair. Trust opens up the possibility of trustworthiness. Without trust, trustworthiness is impossible.

*2.2. Domain Terminologies*

In addition to this complementary relationship, Flores and Solomon further classified trust into five levels or degrees, namely **simple** trust, **basic** trust, **blind** trust**, authentic** trust, and **articulate** trust. Simple trust is a kind of naïve, unchallenged, unquestioned, and unarticulated belief, such as children trust their parents. This kind of trust is without distrust. "The absence of distrust is what makes it so simple." [15] **Basic trust** is a set of baseline norms that take for granted, such as a guarantee of personal safety and privacy in a peaceful society. Blind trust is a sort of stubborn mentality or self-deception mindset. **Authentic trust** holds trust and distrust in balance. People who have authentic trust understand its risks and vulnerabilities. **Articulate trust** is spelt out as a contract or agreement with enforcement mechanisms in place.

From a measurement perspective**, the articulate trust** provides fairness, transparency, and generalizability**,** which can be associated with **a justice class of dimensions**. At the next level below, **Authentic trust** concerns risks and long-term accountability, which we can consider **a security class**. In other words, we require a TAI system to quantify threats and potential benefits and provide a balance option [25]. **Basic trust** is frequently ignored and left unspoken, known as unarticulate trust. For example, the result of a strategic decision produced by AI/ML can be at least reproducible without explicitly mentioning it. To some extent, "basic trust" can be loosely associated with the **dependence class** (Refer to Figure 2). Although there is a loose relation between a domain with a dimension class, each dimension can align with a different trust domain. It depends on the objectives. For example, if we want to make a strategic decision for a public investment project (e.g., public transport), Fairness (justice category), accountability (security category), and sustainability (dependence category) must be considered. This example gives rise to dimension concepts.

*2.3. Dimension Terminologies*

Figure 2 shows ten-dimensional concepts. We can roughly classify them into three categories: justice, security, and dependence. Under **justice**, there is the concept of **transparency** and **explainability** dimension. The word **transparency** means a substance has the property that allows light to pass through it, and the objects behind it can be distinctly seen. It is opposite to opacity or a black box. From an AI perspective, transparency has become one of the mandatory norms required by the General Data Protection Regulation (GDPR) [26] because the decision-making process assisted by AI exceeds human expert capability [27]. Transparency can be implemented by algorithmic transparency[28]. Wortham argued [29] that transparency is often intertwined with ideas of accountability and responsibility at the objective level. However, the closest definition of transparency is explainable[30]. The definition of "**explainable**" is understandable and interpretable and has reasons for TAI. In other words, something or a story is justifiable and acceptable. It is the state of being fair.

**Fairness** is an essential part of a justice class. Diversity and non-discrimination are complementary to fairness. **Diversity** is the condition of having or being composed of different elements. One of the advantages of diversity is "The Wisdom of Crowds." [31] With enough diversity of data, the machine can learn or formulate unbias rules. Puri et al.[32] adopted a considerable scale and diverse dataset to accelerate the algorithmic advancement, which they called AI for code. Diversity aims to formulate wisdom and mitigate our ignorance or discrimination. **Non-discrimination** means being on an equal footing. Veale and Binns [33] argued that algorithmic and learning models could produce biases if the machine learns from historical data, such as gender, ethnicity, sexuality or disability. Their study offers three potential models to mitigate possible discrimination. Similarly, Wachter et al. [34] addressed this issue from a legal perspective. They intend to bridge the gap between non-discrimination law and AI algorithms known as "contextual equality."

The concept of non-discrimination brings another closely related concept, namely **Generalizability.** It means the degree to which we can apply our training results to a broader context. Paul et al. [35] proposed a novel method for AI fairness and generalization, known as the training and representation alteration (TARA) method. However, generalization often encounters the issue of preserved privacy.[36]

We classify **privacy** as a part of **security**. It is the state of not being shared by others—privacy matters for many reasons, such as ensuring self-determination, safety, and trust. Privacy is vital when we use AI/ML for medical applications. Kassis et al.[37] proposed a next-generation method known as "secure, privacy-preserving and federated ML" that can "protect patient privacy and promote scientific research on large datasets that aims to improve patient care." This concept is closely related to data governance and ownership.

**Data governance** guarantees that the collected data is the safety, quality, integrity, and usability for authorized users and stakeholders during the data's life cycle, which is until the data is destroyed or archived [38]. Kroll offered a set of best practices by using data in automated decision-making. When we encounter automated decision-making, the issue of safety becomes critical. "The IEEE is running several standard processes surrounding AI ethics and safety." [39]. **AI safety** is a system issue because "many new AI applications across various society domains and public infrastructures come with new hazards, which result in new forms of harm" [40]. The author argued that there is an accountability gap for AI-driven innovations because people often misinterpret that they could offload their **accountability** from humans to the AI agent due to automatic decision-making.

Based on the OECD organization's AI principles [41], **accountability** "refers to the expectation that organizations or individuals will ensure the proper functioning, throughout their lifecycle, of the AI systems that they design, develop, operate or deploy, in accordance with their roles and applicable regulatory frameworks, and for demonstrating this through their actions and decision-making process." Loi and Spielkamp [42] further quantified accountability as a relational condition that includes at least three elements: responsibility, answerability, and sanctionability. These elements give rise to the class of **dependency**.

One part of dependency is reproducible, replicable, and traceable. It implies that the experimental result should be **reproducible**. However, Halbe-Kains et al. [43] identified some issues of a lack of details in the methods and algorithm code in the application of breast cancer screening, which demands reducing false positive cases. The authors argued that computational reproducibility is indispensable for high-quality AI applications. High quality can sometimes be interpreted as **reliability** and **robustness. Reliability** implies that AI systems meet the expected standard of performance. In other words, the quality of AI systems is dependable, trustworthy, or performing consistently well. **Robustness**, on the other hand, refers to the ability of AI systems to operate without failure under a wide range of possible conditions, including unexpected or adverse circumstances. In computer science, robustness means AI systems can cope with errors when it executes AI codes. Bhagoji et al. [44] introduce data transformations, including dimension reduction, to enhance the robustness of ML. Li [45] argued that non-robust explanations could result in a moral hazard from the perspective of explaining the real patterns in the world. In short, **reliability** is concerned with consistent performance under known, standard conditions, while **robustness** involves maintaining a normal function under a broad range of possible scenarios, including unexpected conditions.

## 3. Details of TAI Taxonomy

The notions of TAI at the dimensional level are highly intertwined. It is very challenging to carve its nature at its joints precisely. These dimensional concepts are more like a web of the network rather than isolated objects. Despite many challenges, we can classify ten dimensions with 33 implementation models and 106 schemes/tools to formulate a TAI taxonomy in detail from a strategic decision-making perspective (See Figure 5: Details of The TAI Taxonomy)

The dotted mesh between domains and dimensions indicates we can select any dimensions, implementation models, and schemes to form a required domain. No restrictions prevent us from picking up a dimension for TAI objectives. For future computational purposes, we can use various dimensions to form three TAI domain vectors ($T_{L=ar}=(D1, D2, ...Dn)$). Each dimension class is just a placeholder for taxonomy. The taxonomy illustrates five levels of categories, but implementation models and schemes are the technical details, and we do not include all these details in the high-level framework presentation (Refer to Figure 2.), which has only the top three layers.

Although dimensions, implementation models, and schemes can be freely chosen, it does not mean that we should treat these items as a big hodgepodge. It would be inconvenient for implementation. Consequently, we classify these items into three classes based on their characteristics: Justice, Security, and Dependence. According to these classes, we can proceed with the survey of TAI

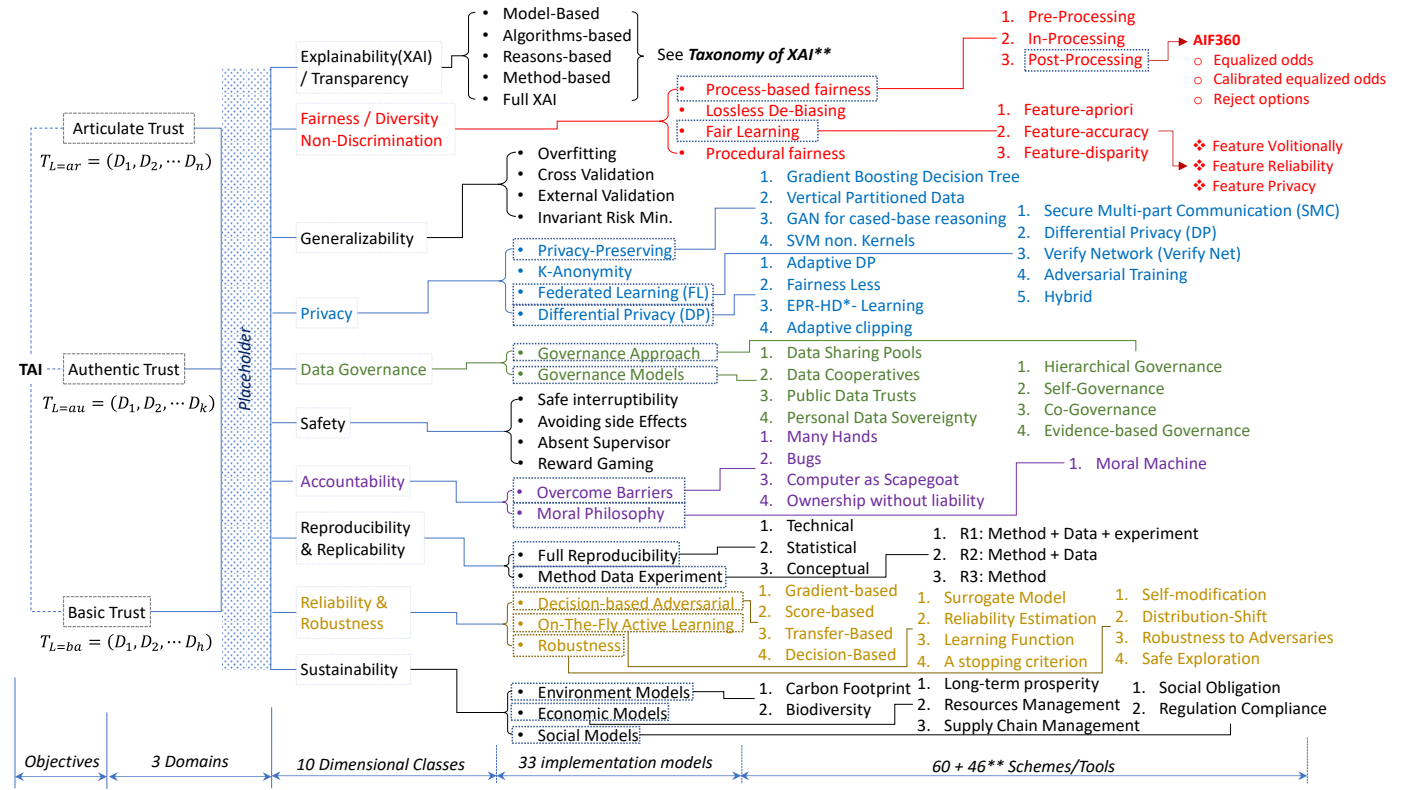

Figure 5 Details of TAI Taxonomy

## 4. Survey on TAI

Like the terminology of trust, the survey is also organised into three levels: objectives, domains, and dimensions. These levels provide a comprehensive view of how AI systems produce strategic decision-making outcomes. The survey represents the overall scope of trustworthy AI regarding different levels of trust under different circumstances and by different value propositions.

*4.1. TAI On Objectives*

Reaching the strategic goal or objectives is often through practical ways of constructing an effective framework. Floridi and Cowls [47] noticed that the sheer volume of proposed TAI principles is overwhelming and confusing. They warned that it could end up with the "market for principles", "where [business] stakeholders may be tempted to 'shop' for the most appealing ones." That is why Benkler [48] urged, "Do not let industry write the rules for AI." He believed that business is running a campaign to bend research and regulation for their benefit."

Floridi and Cowls proposed a four-step process to analyze 47 Ethical AI principles. These steps are Step 1. To report the highest profile sets of ethical principles for AI, Step 2. To assess these principles, whether convergent or divergent, Step 3. To identify the overarching framework, Step 4. To explain how the framework creates subsequent laws, rules, regulations, technical standards, and the best practices for TAI. Their analysis results in the unified framework defined by five overarching principles of Ethical AI: Beneficence, non-maleficence, autonomy, justice, and explicability. Four ethical principles were drawn from the "Georgetown mantra" [49] for bioethics in 1978. The origin of the Georgetown mantra can be traced back to J.F. Kenndey's landmark address about the rights of consumers in 1962. Figure 6 shows the framework from a historical perspective.

Compared to Floridi and Cowls's five principles framework (See Figure 7), the British government proposed a simpler data ethics framework with only three overarching principles published in 2018 [50]: Accountability, Fairness, and Transparency. **Accountability** stands for "the public, or its representatives are able to exercise effective oversight and control over the decisions and actions taken by the government and its officials, …." **Fairness** means eliminating unintended discriminatory effects on individuals and social groups. **Transparency** implies actions, processes and data are made open to inspection by publishing information about a decision model in a complete, open, understandable, easily accessible, and free format.

With respect to fairness, Altman et al.[51] proposed a harm-reduction framework for algorithmic decision-making fairness. The authors claimed that "an algorithmic decision is unfair because the algorithmic decision rules trained by one dataset cannot automatically apply to some individuals who represent the dataset disproportionally. It can result in unintended and even unforeseen harmful effects and contribute to systematic inequality of opportunities." Subsequently, the authors suggested four elements of any algorithmic decision and formulated a fishbone framework. It is mainly applied to the judiciary software known as the Northpointe

Correctional Offender Management Profiling for Alternative Sanctions(COMPAS), which is a type of decision tree. The proposed framework adopts four different lenses of counterfactuals for decision algorithms[2] for the potential outcomes at some critical decision points. The authors concluded that algorithmic fairness must consider the foreseeable effects on the individual's well-being during the algorithmic decision, implementation, and practice.

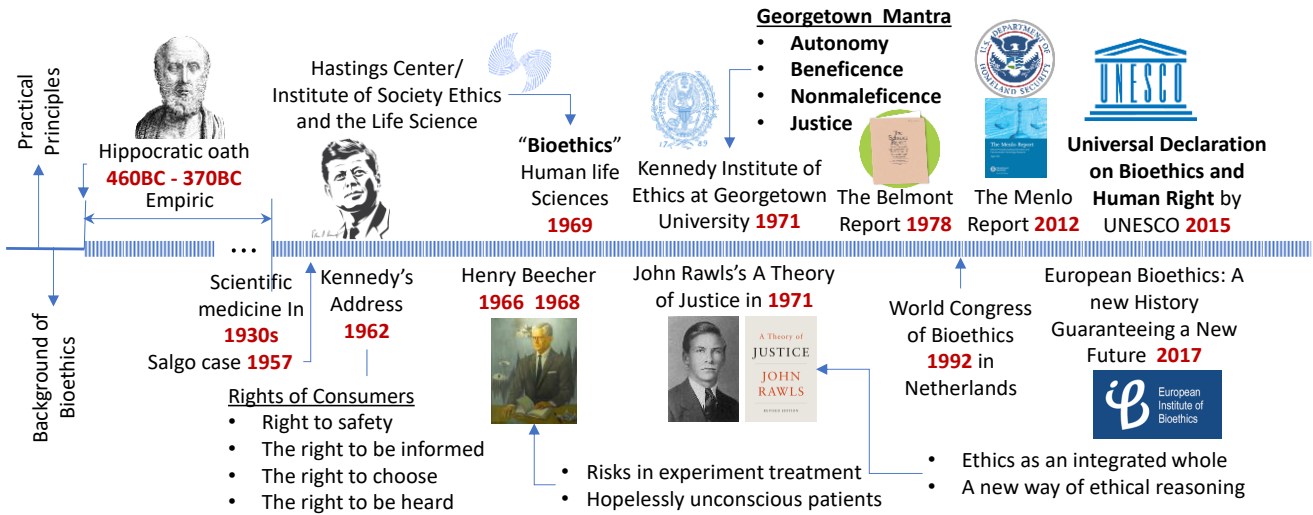

Figure 6 A Brief History Perspective of Ethical AI

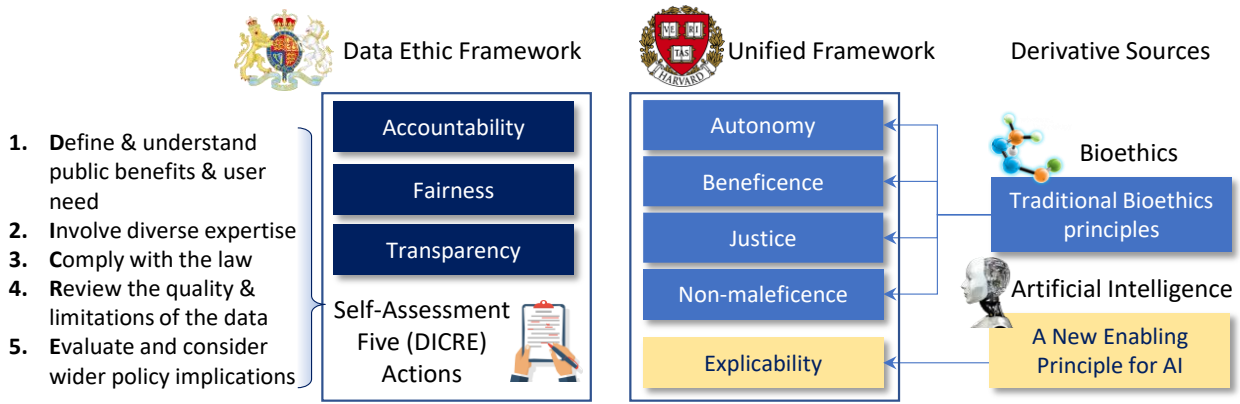

Figure 7 Two Ethic Framework Comparison

In contrast to a particular application, the Future of Life Institute (FLI) contributed ten comments and suggestions on the AI risk management framework (AI RMF) by responding to the National Institute of Standard and Technology (NIST) ' ten requests[3]. However, the AI RMF is less organized due to the randomness of commentaries and suggestions.

Landers and Behrend [52] introduce a framework for evaluating fairness and bias in high-stakes AI predictive models for decision-making from a psychological perspective. They formulated an evaluating framework with 12 crucial components classified into three categories: a.) AI model components include model design, development, features and processes, input data and output results, b.) algorithmic components, and c.) meta or goal-oriented components. These components raise some critical questions for strategic decision-making, such as who should audit the auditors.

By comparison with a component-based TAI framework, Choudhury [53] proposed an ecologically valid framework. The proposed framework is built upon human factors models. The author intends to fill the gap in understanding the dynamic and complex interaction between clinicians and TAI for health care management and medical practitioners. The novelty of the conceptual framework is to consider TAI as a part of the ecological environment, including different interactions between a patient and a medical practitioner. The author argued that the factors shaping people's decisions on TAI could be divided into three levels: governance, organization, and individuals due to the complexity of healthcare systems. Although the proposal sounds quite promising, the model still needs some practical cases to support the abstract idea.

Instead of narrative principles, Machine Intelligence Garage's (MIG) Ethics Committee [54] launched a series of questions for the ethics framework because they argued that questions could illuminate a place where principles should be considered in practice rather

---

[2] i.e., what if: a.) not based on race, b.) not including privacy information, c.) not used algorithm at all, d.) using a better algorithm

[3] For example a.) how to manage a massive scale deployment with the aggregate risks from low probability and high consequence effects (e.g., earthquake impacts on a massive scale of the self-driving car deployment, a recommended algorithms that recommend extreme contents could lead to radical activities or social unrest), b.) how to proactively ensure the alignment of evermore powerful general AI systems

than impose some universal "correct" answers. The framework consists of seven concepts[4] corresponding to 71 questions. Among them, two concepts' have over 50% (35) questions. Although MIG asked some excellent questions regarding TAI, the framework can be more beneficial if the framework is better organized. It is unclear whether these concepts and the associated questions are equally important in practice. To understand the importance of each principle at different levels, we should focus on TAI domains.

*4.2. TAI On Three Domains*

At the domain level, we have articulate, authentic, and basic domains for different levels of trust. The connotation of domain trust has the meaning of domain knowledge. Without domain knowledge, it would be impossible to articulate or authenticate any trust. We proceed with the survey based on domains and associated dimensional classes.

*4.2.1. TAI On Articulate Trust*

Articulate trust is the trust that has been "spelt out" as a kind of belief. It is beyond simple trust and knows the risks and vulnerabilities well.[15] The articulate trust at least requires measurements of explanation, transparency, fairness, and non-discrimination because it needs "spelling out". We can also add other dimensions or measurements to consolidate the articulate trust. It depends on the context of a particular AI system. People will not trust AI systems if we can not articulate how AI training algorithms or models make decisions transparently, fairly, and non-discriminately. Bennet and Keyes' work [55] provides two concrete cases (autistic and vision-impaired patients) to address the unfairness issue regarding AI training algorithms for disabled people because the algorithmic or statistical learning of unfairness is a disparity due to the collected dataset being biased. The authors suggest that AI ethics should move beyond simple or algorithmic fairness toward the notions of justice. The idea of justice [56] is considered a virtue of the social institution. It formulates the moral fabric of modern societies and civilizations. The paper concluded that simple algorithmic fairness is unfair.

*4.2.1.1. Explainability and Transparency*

By contrast, Sharma et al.[57] adopt a counterfactual method to explain fairness, transparency and robustness. By principles, it depends on how many input values can be changed. If a person receives an undesired(i.e., positive) outcome, the method changes input values to achieve desirable(i.e. negative) output. Thus, it is known as the counterfactual method. The authors create a model called Counterfactual Explanations for the Robustness, Transparency, Interpretability and Fairness of AI (CERTFAI). The model is built upon Wachter et al.'s work [58]. Sharma et al. [57] add a genetic algorithm to select input values randomly plus Counterfactual Explanations Robustness (CER) score, which is a fitness value between feasible generated points and an input instance. The paper claims that Wachter's counterfactual model could not handle categorical data. Although counterfactuals have become a popular method for explainability, Byrne [59] asserted that "not all counterfactuals are equally helpful in assisting human comprehension." One reason is that many AI models involve complex algorithms that generate predictions based on large amount f data and patterns that are not readily apparent to humans. One should look beyond a simple counterfactual method to effectively attain explainable AI (XAI).

During the last decade, numerous ways of XAI have been developed because understanding and interpreting AI can be derived from different perspectives. It depends on users[60], logic[61][62], biases[63], algorithms[64], responsibilities[65], methods/ processes, models [66] [67], systems [68], stage[69], costs, and reasons[70]. Some researchers suggested that we should explain from a social science perspective[71]. Others [72] argue that it is not necessary to explain but interpret it. Burns et al. [73] proposed interpreting AI through hypothesis testing. However, Gilpin et al. [74] disagreed and argued that the interpretation is insufficient. Whether we should explain or interpret it, we should look at how these various ways relate to each other and how we can apply it with a particular application to a particular case regarding trust level or TAI objectives. Subsequently, we summarize various types of XAI into five categories (See Figure 8): model-based, algorithm-based, full XAI, reason-based, and method/process-based XAI.

The most compelling type of XAI appears to be the full XAI because it matches Marr's three levels of analysis or explanation architecture. Gilpin et al.[74] classify XAI into three categories: presentation, processing, and explanation. They intend to explain the questions of "why" and "why should" rather than interpreting AI/ML in comparison with Doshi-Velez and Kim's classification [75], which also has three layers. One of the potential issues of the "why" model is that one "why" could generate numerous subsequent "why"s. As Immanuel Kant indicated, "The principle of question propagation: 'every answer given on the principle of experience begets a fresh question.'" [76] The interpretation model is much more effective because some ethical values are beyond the reasons.

Gilpin et al. [74] highlight three ethical dilemmas for the challenge of XAI: "interpretable and explainable", "persuasive and fact-based transparency", and "manipulation and explanation". If we consider "interpretable" as a kind of predictability, then the more effective interpretability, the less explainability. Conversely, the more explainability, the less interpretability. Similarly, Herman [77] also indicated that human evaluations create persuasive systems rather than transparency because "human evaluations imply a strong and specific bias towards simpler descriptions." Herman presented the ethical dilemma between manipulation and explanation by asking questions about how to draw a line between unethical manipulation and ethical explanation or how to balance our own interests in transparency and ethics with our desire for interpretability. These dilemmas are very challenging. Consequently, we propose "what, what and how" instead of "why, why and why" to draw a line between interpretation and manipulation because we can clarify "What is a purpose?" at the objective layer, "What is a process?" at the domain layer, and "how can we implement the process?" at the dimension layer. (See Figure 9) These questions can provide a full explanation of AI (XAI) to underpin TAI.

---

[4] These concepts are 1.) Be clear about benefits (10 Q), 2.) Know and manage the risks (11Q), 3.) Use data responsibly (18Q), 4.) Be worthy of trust (17Q), 5.) Promote diversity, equality and inclusion (8Q), 6.) Be Open and understanding in communication (7Q), 7.) Consider the business model (8Q).

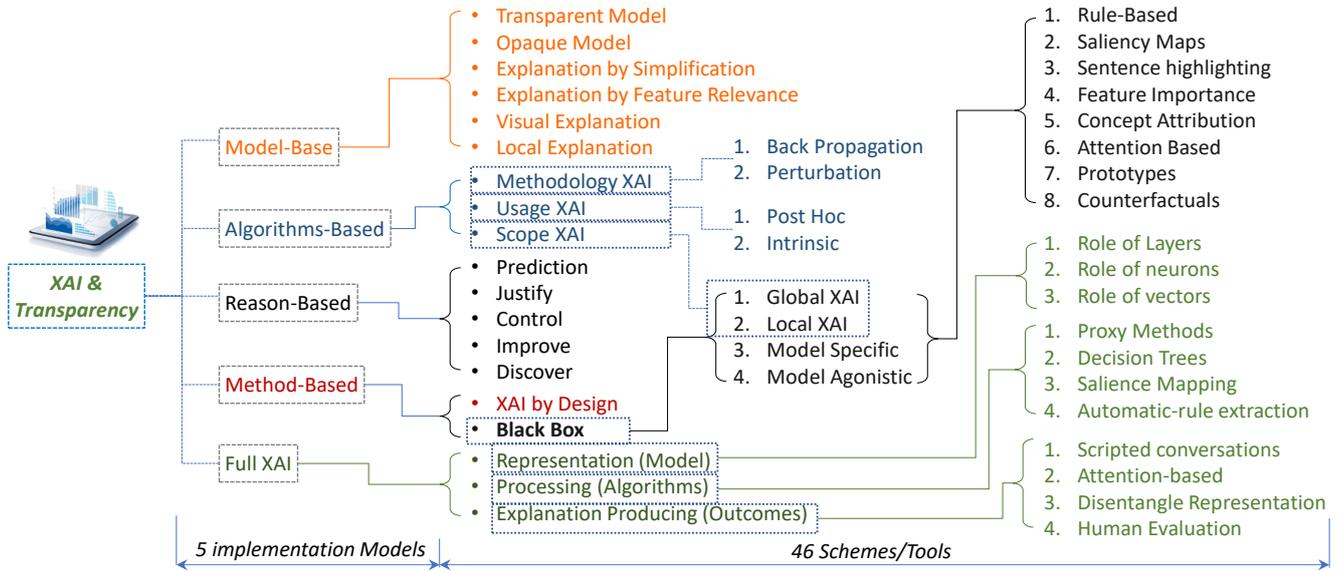

Figure 8: Taxonomy of XAI

In addition, one of the recent developments is that OpenAI announced [78] a novel way for XAI, which uses AI (GPT-4) to explain AI (GPT-2) recently. We can consider using a black box to explain a black box. At a high level, the OpenAI XAI of AI is a three-step process: explain, simulate, and score. The principle of XAI of AI uses the subject model to hypothesize the model behaviour. Based on the hypothesis, the process applies a simulator model to make a prediction and then uses the score to evaluate the hypothesis prediction compared with the actual prediction value. However, not every neuron can be explained because some neurons may correspond to polysemantic concepts. XAI always consists of different dilemmas.

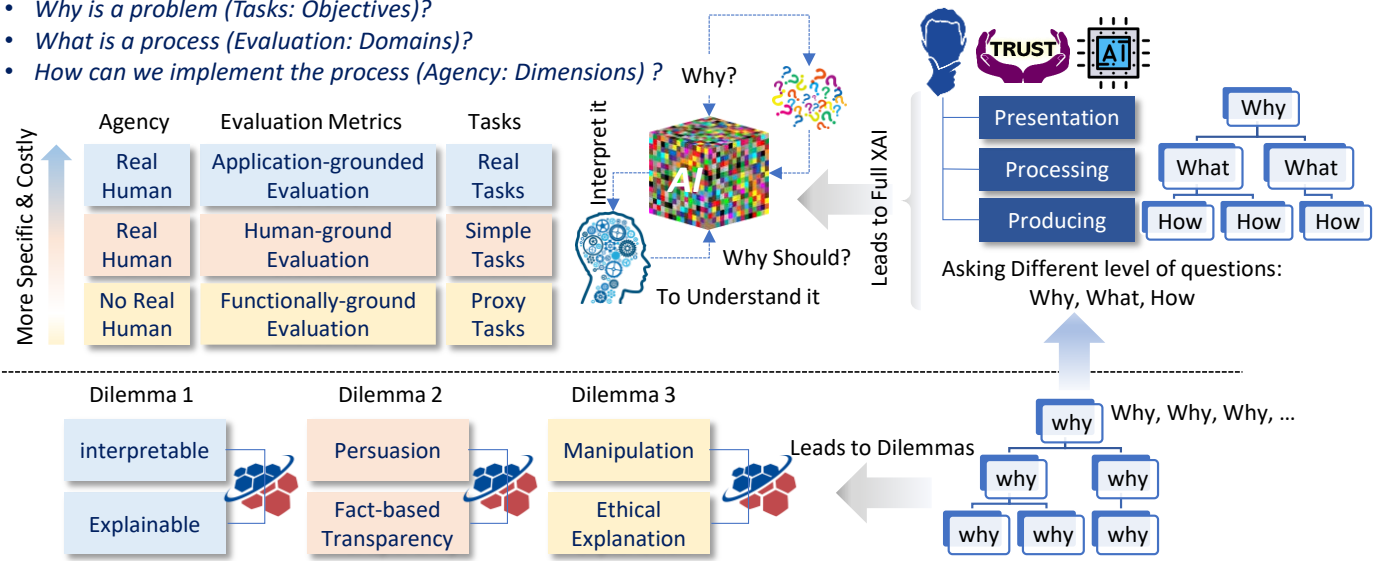

Figure 9: Asking Different Questions for Full XAI

*4.2.1.2. Fairness and Diversity*

Considering these dilemmas from a broader justice perspective, these issues are closely related to fairness and diversity. The concept of fairness and diversity can be broadly categorized into three main categories: a.) process-based fairness [79] [80], b.) Lossless De-biasing fairness [81], and c.) fair learning [83], and procedural fairness [82] (Refer to Figure 5)

According to [79], process-based fairness ensures that an automotive decision is less biased than human. However, extensive evidence shows that many AI biases are inevitable because human social biases are embedded in training datasets and could be amplified through the ML development process and complex feedback loops. The study proposes pipeline process algorithms, known as pre-processing algorithms, to mitigate these biases. One of the open-source tools is AI Fairness 360 (AIF360) [84] [84]. It offers four pre-processing algorithms: a.) **Re-weighing** pre-processing, b.) **Optimized** pre-processing, c.) **learning fair representations**, and d.) **disparate-impact remover**. The last two algorithms are ideal tools for processing transparency. AIF360 also provides three **in-processing algorithms** to eliminate unwanted biases, including a.) Adversarial debiasing, b.) Prejudice remover, and c.) Meta fair classifier. Finally, if we can only access the output predictions, AIF360 provides three **post-processing algorithms**: **a.) Equalized odds, b.) Calibrated equalized odds, and c.) Reject options** classification. The advantage of AIF360 provides tangible schemes and tools to implement fairness at the practical level. However, it is not a silver bullet. The metrics of AIF 360 can be considered distributive justice. It will not capture every aspect of fairness in any circumstance. Sometimes, the removed or equalized odds criteria might trigger valuable information loss and decrease the accuracy of the prediction.

Zhou et al. [81] present an information lossless solution that can maintain fairness or de-biasing. The basic idea of their solution is to eliminate the oversampling problem for favourable representation while undersampling for a minority group. In other words, they oversample the underrepresented population to balance the majority and minority of the demographic population. Although this solution preserves the content in the original dataset, the predicted accuracy is sacrificed by oversampling the underrepresented population. The most famous case is the US presidential poll by The Literary Digest in 1936 [85] [86], which is the failure of oversampling.

Another fairness proposition is ideal fairness, which is basically driven by the decision outcome or distributive fairness (ends). Grgic-Hlaca et al. [82] propose procedural fairness or fair learning (means) to consider the input features and moral judgments. The solution introduces three new scalar measurements of procedural fairness: feature apriori, feature accuracy, and feature disparity by considering feature volitionality (i.e. "done by personal choice"), feature reliability, and feature privacy. Subsequently, this approach avoids a situation of binary discrimination. The authors argue that they can achieve procedural fairness with little cost compared to other fairness algorithms. However, some loss of accuracy is inevitable. Furthermore, the solution does not hold in generalizability.

*4.2.1.3. Generalizability*

Generalizability is closely associated with the notion of reproducibility. McDermott et al. [87] propose a taxonomy of reproducibility (more details in section 4.2.3.1), including technical, statistical and conceptual components. However, Azad et al. [88] highlight some common errors that lead to poor reproducibility. The paper suggests that it is necessary to ensure generalizable models' performance before such models are deployed on a large scale, especially for medical applications. The authors propose cross-validation, overfitting, and external validation techniques to increase generalizability in corresponding to technical, statistical, and conceptual components. Arjovsky et al. [89] propose invariant risk minimization (IRM) algorithm to extract share features from multiple training environments and to enable out-of-distribution (OOD) generalization. If we consider IRM as an objective function at the algorithmic level for the goal of Invariance Across Environment (IAE), then Kartik et al. [90] provide the IRM-game to optimize a specific IRM at an implementation level.

The advantage of the IRM is that it does not have to know the causal model, which is often very challenging to be specified. It also does not have to know every bias. Instead, IRM selects an appreciated environment and partitions the data into disjoint environments. However, Kamath et al.[91] argued that IRM can fail to capture "natural" invariances, at least when we apply it to the linear form. Consequently, it can lead to a model of poor generalizability in new environments. Furthermore, even if IRM can capture the "right" invariances, the paper shows that IRM only learns a sub-optimal predictor due to the loss function not being invariant across environments.

In a nutshell, generalizability allows you to replace data values (training dataset) with less precise values (new environment) regarding various learning techniques but still capture various actual patterns or realities. However, on the other hand, if the data values can still preserve certain information, it might trigger the re-identification of individuals or reveal private information unintentionally and make TAI insecure. This issue leads to the topic of authentic trust corresponding to the security class.

*4.2.2. TAI On Authentic Trust*

Authentic trust implies what is exactly claimed according to fact. Unlike simple trust, authentic trust does not deny distrust but accepts it, transcends it, absorbs it, and overcomes it, just like security issues, such as hacking, ransomware, password attacks, and malware. Authentic trust aims to solve the "distrust" due to cybersecurity issues. Under the security class, we have four dimensions (See Figure 5): privacy, data governance, safety, and accountability. Privacy and data governance can be considered as one coin of two sides. However, we separate privacy and data governance because of measurements.

Montenegro et al.[92] propose a solution known as a privacy-preserving general adversarial network(GAN) for case-based reasoning (CBR) for medical image analysis applications. The authors argued that many current privacy-preserving methods have three drawbacks regarding realism, privacy, and explanation power. Realism drawbacks imply issues for the traditional CBR method, such as the K-Nearest Neighbour (K-NN) algorithm. The primary difference between CBR and other rule-based reasoning is that CBR retrieves the most relevant cases from memory and adapts them to fit new situations [93]. Privacy and explanation drawbacks refer to a dilemma of either failure to preserve relevant semantic features or failure to ensure privacy. Therefore, the paper develops a privacy-preserving GAN model to privatize CBR in a clinical setting. Moreover, the study adds counterfactual explanations to enhance the model's explanatory power. However, it does not matter how sophisticated it is. The model only remains as a CBR. The critical issue of CBR is inadequate for large and highly structured case bases.[94]

*4.2.2.1. Privacy*

Yu et al. [95] propose privacy-preserving support vector machines (PP-SVM) by leveraging nonlinear kernels for the problem of distributed privacy-preserving data. The solution is to establish the global SVM classification model from decentralized data in multiple locations without compromising the data's privacy. However, if there is an intermediator between stored data and the global SVM solution would not work. Moreover, they assume that data is partitioned horizontally and that the same features are collected for different objects. In contrast, Karr et al.[96] proposed a solution that can execute secure regressions on vertically partitioned data, which means the datasets can have identical records with different sets of features. They claim their solution can manipulate a secure matrix "by pairs of owners to compute off-diagonal blocks for the full data covariance matrix.". The advantages of vertically partitioned data are 1.) It can separate relatively slow-moving features stored in cache memory, 2.) It can provide extra flexibility for different levels of privacy (e.g. passwords or personal information), 3.) It can reduce the amount of concurrent access. 4.) It works well with column-oriented databases like HBase or Cassandra. The disadvantages are 1.) It is relatively challenging to join various datasets, 2.) It requires more joins to retrieve data.

The Gradient Boosting Decision Tree (GBDT) [97] is another valuable learning technique for privacy to optimize the predictive value (to minimize the loss function) through an iterative learning process, especially for classification and regression tasks. Li et al. [98] propose a privacy-preserving algorithm for the new GBDT training algorithm known as DPBoost. The basic idea is to split the value (or threshold level) of privacy budgets (the maximum tolerance is allowed to each user for revealing information) into serial and parallel. However, the DPBoost algorithm is too complicated and costly. Compared with the baseline GBDT, the training time for a meaningful result is significantly high, up to 3.62 times (average 2.33). Furthermore, the paper does not show the trade-off or comparison between the privacy budgets and the model's performance. It is challenging to compare the privacy budget with the proposed algorithm's performance (accuracy of the prediction rate).

Another traditional approach for privacy is "k-anonymity" [99], which is to generalize or suppress a dataset's attributes (columns) until each row is at least the same as k-1 other rows. The basic idea of k-anonymity is "hiding a person in a crowd" to reduce the risk of the person being re-identified by indirect identifiers. The method is often applied to the healthcare and banking industries. However, Aggarwal [100] showed that it is unsuitable for high-dimensional datasets.

Alternatively, researchers leverage federated learning (FL) or collaborative learning to preserve personal privacy and data security locally. The lexical meaning of federation refers to a type of organization relationship in which each joined member maintains a certain degree of independence or autonomy for the local affairs. FL trains decision rules (or algorithms) across multiple local servers that keep their local datasets instead of sharing with others or uploading them to a central server. According to Mothukuri et al. [101], FL is one of the favourable methods where security and privacy become primary concerns because FL can minimize the risk of privacy leakage. The authors highlight five major state-of-the-art algorithms to mitigate identified threats and enhance general privacy-preserving features: a.) Secure Multi-part Communication (SMC), b.) Differential Privacy (DP), c.) VerifyNet, d.) Hybrid, and e.) Adversarial training. (See Figure 5)

**SMC** is an FL framework in which multi-participants contribute their inputs for a joint training model. The main advantage of SMC is that it only needs to encrypt parameters rather than the entire input dataset. However, the cost of the SMC solution is its performance because it often needs more time to train a model. The challenge is how to design a lightweight SMC solution for FL clients. **DP** solution is similar to k-anonymity, which adds noise to personal attributes. The advantage of DP is to solve the dilemma between predicting accuracy and concealing private information. Researchers proposed many solutions for the dilemma, such as adaptive DP [102], Fairness lens [103], Efficient Private Empirical (EPR) minimization for high-dimensional learning [104], and adaptive clipping [105]. Overall, there are three strategies: 1.)adding output perturbation[106], 2.)differentially private stochastic gradient descent [107], and 3.)optimizing the privacy budget [98] [107].

The primary issue of DP is that the statistical data quality is degraded. Within a training FL environment, we only add the noise to each client's upload parameters. Although **DP-FL** offers a good solution for privacy protection, it also adds uncertainty to upload parameters. This issue could cause the overall poor performance of a training model.

**VerifyNet** [108] is another FL framework that proposes a double-masking protocol to verify results between clients to a central server. Although the idea of the double-masking protocol seems to be a pursuable solution, the only issue is the scalability because the central server has to verify each client. It could overload the central server if there are many clients. In terms of scalability, Truex et al. [109] propose a **hybrid** model for FL. The paper presents an alternative approach to address the vulnerability for inference and DP's poor performance. The solution is to combine both DP and SMC for FL. Therefore, they call it a hybrid. Similarly, Teng and Du [110] also presented a hybrid model for vertical partition data mining. The advantage of the hybrid model is that it can balance privacy and prediction accuracy (training model performance). However, Trues et al.'s [109] experiment results are only based on eight features or dimensions. The question is whether this approach will still maintain its performance for the high-dimensional dataset is unclear.

Adversarial Training (AT) or Ensemble AT(EAT) [111] is another strategy to improve privacy for FL. The idea of AT is straightforward: injecting noise or perturbations into the dataset to increase the robustness of the training model. EAT is an enhancement version of AT. The EAT model reduces the vulnerability but stops short of increasing the model's accuracy when facing elaborate black-box attacks.

*4.2.2.2. Data Governance*

On the other side of the coin, Data Governance is a function or process of data management to ensure the data quality, integrity, security, and usability of data collected by any organization [112]. It is an ongoing process from the starting point of data being collected to the end point of data being destroyed. During this lifecycle, data governance must guarantee 1.) data's accessibility to permitted users, 2.) auditable, and 3.) compliance with regulations. Due to the prevalence of AI/ML, Gritsenko and Wood [113] highlighted that there is "a shift towards a special form of design-based governance with power exercised ex-ante via choice architectures defined through protocols, requiring lower levels of commitment from governing actors." Kooiman [114] defined three data governance approaches, self-governance, co-governance, and hierarchical governance. **Self-governance** means "taking care of themselves". The advantage of self-governance is to respond to the regulatory vacuum. **Co-governance** implies "where social parties join hands with a common purpose in mind." It requires sharing power, responsibility, and information. The co-governance is a type of meta-governance.[115] This approach might include co-operation, co-management, and collaboration. The main advantages of co-governance are flexibility and scalability. **Hierarchical governance** refers to that governing is superimposed upon those governed. Those governing people adopt a "top-down" method and ask people to be governed to comply with the rules and laws governing people created. The key advantage of hierarchical governance is operational efficiency. The disadvantage is that information can be distorted at each hierarchy level. In addition to design-based governance, we can also have evidence-based governance. De Bruijn and Janssen [116] argued that cybersecurity is "a global phenomenon representing a complex social-technological challenge for the government." They present evidence-based framing strategies to deal with the challenges. This approach has also been widely applied to AI explanation and transparency.

If a governance approach focuses on a broad scope, then a governance model emphasizes detailed implementation. Micheli et al.[117] shed light on four different models of data governance known as data sharing pools (DSPs), data cooperatives (DCs), public data trusts (PDTs) and personal data sovereignty (PDS). Each model has its goal, application, value and principles (Refer to Appendix Table 1)

*4.2.2.3. Safety*

AI safety is one of the components of the security class. It is also a priority issue [118] for TAI because many accidents caused by unintended and harmful behaviour of AI/ML systems emerge from poor, unconscious and imperfect designs. Leike et al. [119] classify these problems into robustness and specification problems from a reinforcement learning perspective. However, based on the phenetic taxonomy method, we classify robustness as one of the dependent dimensions from a basic trust perspective.

The safety dimension consists of four issues with four designing questions: 1.) safe interruptibility, 2.) avoiding side effects, 3.) absent supervisor, and 4.) reward gaming (See Figure 5). **"safe interruptibility"** implies that humans can interrupt a machine or robot's actions at any time. The question is how to design an AI agent that neither pursues nor avoids human interruptions. **"avoiding side effects"** means the AI agent has some unexpected and negative side effects that are not intentionally designed. The question is how to minimize these negative effects, especially those irreversible or difficult to reverse. **"Absent supervisor"** is that the AI agents constantly act with and without human supervision. Our question is how to design an AI system with constant behaviour when supervision is absent. **"Reward gaming"** concerns that an AI agent exploits the system errors in the reward function to get more rewards. Our issue is how to design an AI system to break this ill reward loop. If we consider each above question from a responsibility perspective, it leads to accountability.

*4.2.2.4. Accountability*

Nissenbaum [120] proposed four accountability erosions or barriers: 1.) the problem of many hands, 2.) the problem of bugs, 3.) blaming the computer, and 4.) software ownership without liability due to the ubiquitous delegation in a computerized society. Cooper et al. [121] revisited these barriers and uncovered new challenges under data-driven algorithmic systems. They intend to create a relational accountability framework that can be applied in practice and gradually overcome these barriers.

Building a large and complex AI system could involve many phases: conceptualization, design, development, deployment, and operation. If the system becomes malfunctions and causes harm, it is not easy to pin down a particular component at its source and accountable person. For an ML pipeline process, the many-hands barrier becomes very evident because the ML pipeline is a dynamic process that involves many sub-processes and different groups of people collaborating, such as product designers, software engineers, managers, researchers and data scientists.

The second barrier is "Bugs", which consist of various errors: "modelling, design, and coding errors." Bugs could be both predictable and unpredictable. Unexpected bugs become a barrier to accountability because they are expected and treated as acceptable cases.

The third barrier is "the computer as a scapegoat." No physical machine has its own intentionality because it does not have thoughts, beliefs, desires, and hopes. "If people place responsibility on technology, not its developer, owners, and operators, it reduces accountability to piecemeal."[121] This issue becomes even more complicated for AI/ML because of the black-box nature of the neural net and the limited power of explainability.

The last barrier that Nissenbaum [120] presented is "ownership without liability," The computer industry is evolving towards consolidating property rights while avoiding liability. This phenomenon reinforces the barrier to being accountable. The simple reason is that the liability costs, but the property right brings fortunes. Cooper et al. [121] suggest putting the conditions necessary for a moral and relational accountability framework in place to enforce processes of explanation, transparency, auditing, and robustness. They argue that accountability is moral and relational decisions.

Awad et al. [122] created the moral machine (i.e. an auto-driving car faces a moral decision) experiment to investigate how machines will make moral decisions. Their primary challenge is quantifying social expectations about the ethical principles to regulate the machine's decision behaviour. The experiment generated 40 million decisions in ten languages from millions of people in 233 countries and regions. Their results show that moral preferences are social, economic, and cultural bound. By setting nine scenarios for moral preferences, Awad et al. [122] identify three cultural clusters with different moral choices, although three moral preferences are universal. Moreover, the authors also raise the moral dilemma: if we give a strong preference to children and pregnant women's lives when an accident occurs, how can we explain the rationality of such a decision? These issues give rise to how we can depend on machines to make fundamental decisions. In other words, what is the basic trust of TAI?

*4.2.3. TAI On Basic Trust*

Basic trust is a kind of baseline ingredient without saying. It is a natural understanding of trust. We discuss three dimensions or measurements of basic trust: 1.) reproducibility and replicability, 2.) reliability and robustness, and 3.) Sustainability (See Figure 5). The first dimension consists of full reproducibility and "method data experiment". The reliability concerns decision-based adversarial and on-the-fly active learning. The sustainability dimension consists of the environmental, economic, and social models.

*4.2.3.1. Reproducibility and Replicability*

McDermott et al. [123] identify reproducibility issues in three elements from a healthcare perspective: **Technical Replicability (TR)** means that readers can fully replicate the result claimed by research regarding releasing code and dataset. **Statistical Replicability** (ST) refers to a paper's result that can be replicated under statistically identical conditions. In other words, the result could be different if the reader takes a different set of random seeds or training and test set splits. **Conceptual Replicability** (CR) implies that a paper's result can only be replicated under a conceptually identical condition. CR is closely associated with "external validity," in which a paper's result can be extended to other situations, people, settings and measurements. The authors explained that these components are the replicability criteria because, without TR, the claimed result can not be demonstrated; without ST, the experiment result is impossible to be reproduced if we increase the sample size. We can not deploy the trained result or algorithms in practice without CR.

Gundersen et al. [124] [125] introduced AI reproducibility with six metrics based on boolean combinations of three essential reproducible factors 1.) experiment, 2.) data, and 3.) method (Refer to Figure. 10). **R1** is the most restrictive (equivalent to full reproducible), while **R3** is the most relaxed (equivalent to Statistical reproducible) regarding requirements. Furthermore, the study develops sets of tangible variables as indicators to measure reproducibility. The table of Figure.10 [5] shows the details of collected survey papers from two top conferences (IJCAI and AAAI) in three different years, respectively.

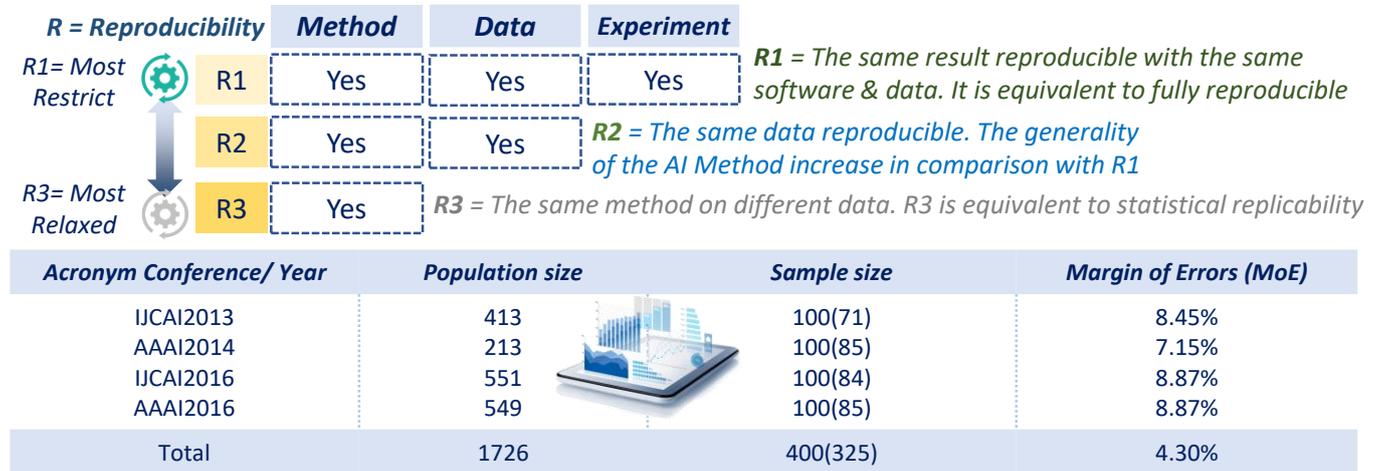

Figure 10: The AI Reproducibility Survey Results for Top AI Journals

Based on the above criteria of measurements, Gundersen et al. showed that no single paper is documented well. The survey concludes with the alarming result that no single paper can be fully reproducible. Although there is a statistical improvement in R1D ("1" means a factor of pseudocode), the rest of the categories (RXD) have no improvement in 2016 in comparison with 2013 and 2014.

Based on the above criteria of measurements, Gundersen et al. showed that no single paper is documented well. The survey concludes with the alarming result that no single paper can be fully reproducible. Although there is a statistical improvement in R1D ("1" means a factor of pseudocode), the rest of the categories (RXD) have no improvement in 2016 in comparison with 2013 and 2014.

*4.2.3.2. Reliability and Robustness*

Saria and Subbaswamy [126] noticed that although ML-driven decision-making systems are increasingly prevalent in some critical safety applications, such as auto-driving, autopilot, or medical diagnosis, we still do not have a framework for reasoning about failures and potentially catastrophic effects.

Brendel et al.[127] address the reliability issue from the perspective of imperceptible and adversarial perturbation inputs for learning algorithms. Adversarial perturbations imply that the slightly perturbed inputs will cause the ML algorithms trained under unperturbed conditions to deliver wrong results. Even worse, this "adversarial perturbation" is ubiquitous and imperceptible to humans. For example, if an auto-driving car (ML algorithms) recognises a degraded stop sign as a 120-kilometre zone, it will cause a catastrophe. Algorithms that intend to increase their ability to counterattack such adversarial perturbations are called "adversarial attacks". The study [127] classifies adversarial attacks into three categories: gradient-based, score-based, and transfer-based attacks. **"Gradient-based adversarial attacks"** simply mean the gradient of loss regarding the inputs. One of the common ways to counterattack gradient-based attacks is to mask the gradients. **"Score-based attacks"** refer to using the model's output or predicted scores to craft adversarial perturbation within dozens of queries. The simple counter measurement increases stochastic elements like dropout into the model, preventing the numerical gradient estimation. **"Transfer-based attacks"** leverage both model information and training data, which adversarial perturbations can be synthesized. Therefore, the attacks are transferrable between models. The defence mechanism is to increase the number of adversarial examples from an ensemble of substitute models. From the above three adversarial attacks, Brendel et al.[127] derived a new category: **"Decision-based attacks"** solely depend on the model's final decision. They claimed three reasons for creating such a category. 1.) Decision-based attacks are highly relevant to many real-world ML systems because the internal decision-making process is unobservable. 2.) The attacks are not dependent on substitute models trained by similar data, 3.) The attacks are much more robust than the other types of attacks. They concluded that "Decision-based attacks" can enhance the robustness of ML models. The issue is how to generate adversarial perturbation to meet the reliable and robust criteria under "Decision-based attacks". This issue gives rise to robustness.

---

[5] The method (M) includes five variables (or X in a factor): 1.) pseudocode, 2.) question, 3.) method, 4.) objective/goal, and 5.) research problem to be solved. The data(D) has four variables: 1.) results, 2.) test, 3.) validation, and 4.) training. Experiment (E) consists of seven variables: 1.) experiment code, 2.) set-up, 3.) software dependencies, 4.) hardware specification, 5.) method code, 6.) prediction, and 7.) hypothesis. With these variables, the authors create three additional metrics, R1D, R2D, and R3D, in which the Results are Dependent (or RXD) on the weighted sum of the truth values of variables.

The robustness consists of four schemes according to [119]: 1.) self-modification, 2.) Distributional shift, 3.) Robustness to adversaries, and 4.) safe exploration. "Self-modification" indicates that the AI system can modify itself to cope with a new environment. The human designer's task is how to build such an AI system that allows it to complete self-modification. **"Distributional shift"** is also known as "domain shift", which refers to the AI system that will adapt well to the deployed environment, not just the training dataset. The question is how to ensure the AI agent behaves robust enough when the deployed environment differs from the training. **"Robustness to adversaries"** signifies that the AI agent can survive adversarial attacks. The job is to ensure that the AI agent can differentiate between friendly and adversarial intentions in a complex environment. The function of **"Safe exploration"** is to enable the AI agent's safety constraints under both normal and initial learning environments.

On the other hand, Moustapha et al. [128] propose a generalized framework for reliability issues with various combinations of four components from an active learning perspective: 1.) a surrogate model, 2.) a reliability estimation algorithm, 3.) a learning function, and 4.) a stopping criterion. The paper claims that the proposed solution supports 39 on-the-fly active learning strategies for 20 reliability benchmark problems. The results of this extensive benchmark can be applied to 12000 reliability problems. The primary idea of active learning is to reduce the cost of the simulation algorithm. By introducing the surrogate model, active learning can replace the expensive evaluation of limit-state function with inexpensive approximation. It saves training costs and energy. The topic of energy saving elicits the board context of sustainability.

*4.2.3.3. Sustainability*

The issue of sustainability is one of the most profound problems for strategic decision-making by enabling AI-based technologies, especially business operations.[129]. Vinuesa et al. [130] address sustainability from a sustainable development goals (SDG) perspective. They classify SDG into three categories, also known as pillars: environment, economy, and society, using a consensus-based expert elicitation process. The authors concluded that the fast pace development of AI technologies needs regulation insight and oversight. Otherwise, AI systems will result in gaps in transparency, safety and ethical issues.

Regarding environmental sustainability, Venues et al. highlighted AI technologies by leveraging large-scale interconnected databases to develop simulation models for strategic decisions on climate actions. "Transactions on Large-Scale Data and Knowledge-Centered Systems" (books series) has published 53 volumes of techniques since 2009. [131] The latest lecture notes [132] provide many recent machine learning techniques to handle large-scale data, including integrated model-based management of time series, neural graph networks, a fairness-aware approach for protected features and imbalanced data, causal consistency in a distributed database, actor model, and elastic horizontal scalability.

These techniques provide different apparatuses for decision-makers to ensure that humans can co-exist on Earth over generations. One of the apparatuses is the carbon footprint measurement. Lacoste et al. [133] propose an approximate way [134] to calculate ML carbon footprints. The calculation includes four inputs: 1.) hardware or server type, 2.) Usage time in an hour, 3.) Cloud Service Provider (e.g. AWS or Azure), and 4.) Region of Computer (a physical location of servers). The output of the calculation is the carbon equivalents (or $CO_{2\text{-eq}}$). The aim of $CO_{2\text{-eq}}$ measurement, instead of $CO_2$ directly, is that other gases, such as methane ($CH_4$), nitrous oxide (or $N_2O$) or even water vapour, also impact the greenhouse effect. The unit of the ML emission is in grams of $CO_{2\text{-eq}}$ per kilowatt hour, which is equivalent to the measurement of a people's transportation(i.e. aeroplane) emission in $CO_{2\text{-eq}}$ per person-km. As the authors indicated, the calculation is just a starting point.

Schwartz et al. [135] argued that the $CO_{2\text{-eq}}$ measurement is "hardware dependent and, as a result, does not allow for a fair comparison between different models developed on different machines.". They argue that the measurement of running time in an hour is highly influenced by other factors, including the underlying hardware, other adjacent jobs running on the same machine, and the number of cores used. Therefore, they suggest measuring floating-point operation (FPO) because FPO has several appealing properties, including 1.) directly computes the amount of work done by the running machine, 2.)FPO is agnostic to the hardware on which the model is run, 3.)FPO is correlated with the running time of the model. However, as the authors indicate, FPO measurement has its limitations because the training model is not only determined by the amount of workload but also by communication between the different components, which the FPO can not measure it. Nonetheless, FPO does demonstrate a comparison between different training models. The study advocates the improvement of state-of-the-art and the model's efficiency, known as "Green AI". Green AI contrasts with Red AI, where researchers only seek to improve accuracy or model performance.

Economic sustainability implies supporting long-term economic growth and minimizing negative impacts on the community's social, environmental and cultural aspects. Making a strategic decision for economic sustainability faces similar challenges of the environment because it requires resource planning, predictive analysis, economics modelling, strategic leadership, and economic competencies. [137] It requires an interdisciplinary [138] and intertemporal approach [136] to query across multiple large-scale databases. The question is how to empower strategic decision-makers using all possible information to make the right and sustainable choice at the right time. Grover [139] proposed the Bayesian Belief Networks (BBN) solution because it can transfer probability science from the objective to the subjective school of thought. The essence of BBN is using inductive reasoning to learn the truth from combining past outcomes with observable events. As we should see, economic and social sustainability are hand in hand. They form a reciprocal relationship.

Lastly, the meaning of social sustainability is "putting people first." It answers the question of fairness and justice in economic environments.[140]. There are some common issues of social sustainability, such as social equality, community resilience, health and well-being, social justice, and human rights. According to "A Theory of Justice"[141], Rawls argued that two principles would be chosen to govern society: the principle of equal basic liberties for all individuals and the principles of distributive justice, which requires that social and economic inequalities be arranged to benefit the least advantaged members of society. Researchers have adopted nearly all possible ML techniques for social science, including social networks, agent-based models, exploring analysis, random forests, neural network and deep learning, text mining, classification and regression trees. It depends on the particular social problem or decisions and datasets, for example, banking decisions regarding financial loans, insurance decisions regarding eligibility, and university decisions regarding student admittance. These social sustainability decisions are sometimes very challenging to determine because we could encounter many paradoxes in achieving the balance between justice and ethics, such as the paradox of tolerance, equality, individualism, and meritocracy. Table 2 in Appendix summarises each TAI scheme of the main contribution, adopted method, and possible areas to be improved.

Overall, in pursuit of a sustainable future encompassing long-term environmental, social, and economic well-being, we may face many paradoxes, including those related to efficiency, growth, choice, and scale. From a strategic decision-making perspective, solving these paradoxes requires a holistic and systemic approach by considering the intertwining of economic, social and environmental systems, balancing operational, tactical, and strategic goals, recognizing trade-offs and unintended consequences, and engaging stakeholders in collaborative decision-making processes. To fully understand a holistic approach, we should clarify concepts of three ethical theories and paradoxes from the overall TAI and multidisciplinary perspective.

## 5. Three Ethic Theories and Paradoxes of TAI

TAI is ultimately a meta-decision underpinned by multi-disciplinaries in a relationship between the trustors (AI users) and trustees (AI Creators) who produce the AI systems based on collected datasets and selected algorithms. Castelfranchi and Falcone [142] suggest that TAI should include computer science, philosophy, social science, economics, psychology, cognition, and behaviour science. If we initially include all the disciplines, the research problem could quickly become convoluted. Consequently, we started with a relatively simple case.

Figure 11 highlights the multidisciplinary landscape of TAI. We have investigated different levels of TAI in detail because we want to know how to implement TAI in practice rather than a narrative description. Computer science or logical algorithms can help us extract all facts and meaningful patterns from our collected datasets. The question is which fact is more important than others. Computer science will not tell us that. Only axiology (theory of value) will unveil the difference. Therefore, reflecting on TAI based on various ethical theories is essential to prioritizing objectives for strategic decision-making when various paradoxical situations occur.

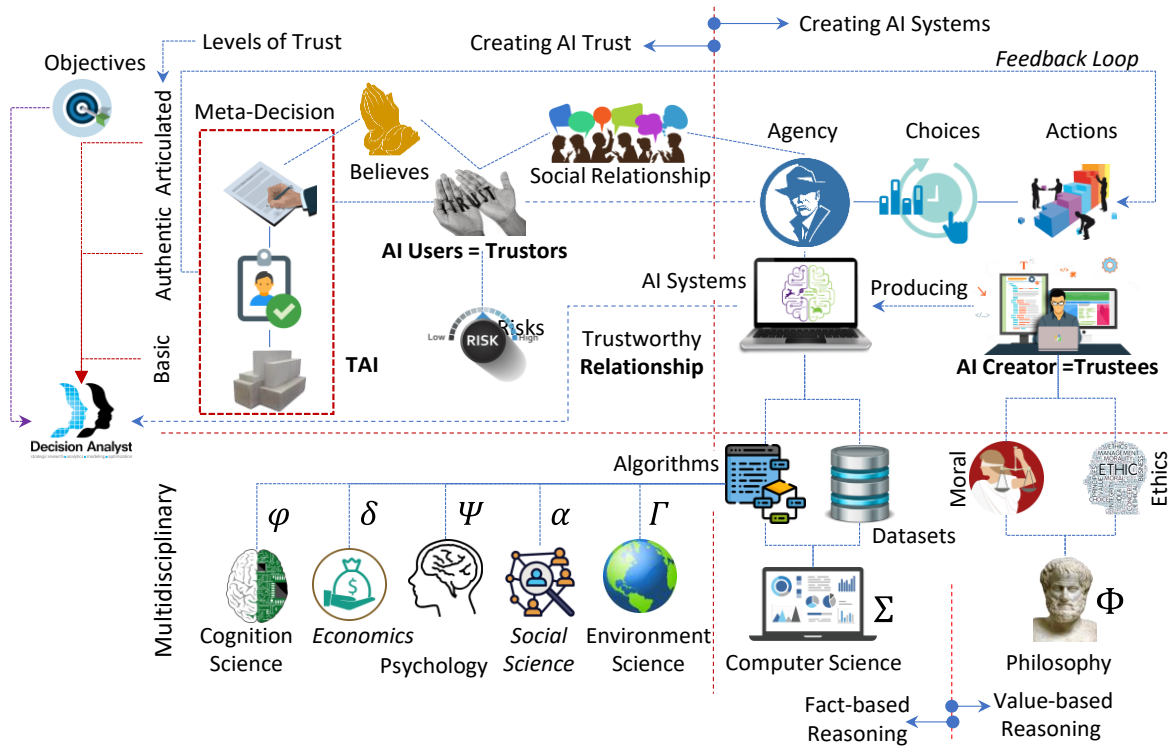

Figure 11: Multidisciplinary Landscape of Creating Trustworthy Artificial Intelligence (TAI)

### 5.1. Three Ethic Theories For TAI

There have been many ethical theories in the history of ethics[6], such as egoism, consequentialism, hedonism, utilitarianism, Kantianism (deontology), and naturalism [143]. Practically, we could encounter three fundamental ethical theories (utility-based, duty-based, and natural law-based) to identify good or bad, right or wrong, and moral or immoral in our decision-making processes.

#### 5.1.1. Moral Utility Theory

Utilitarian ethics is one of the most common theories widely practised today. It was originated by Jeremy Bentham [144] ("the greatest good for the greatest number"), elaborated by John Stuart Mill [145] ("It is better to be Socrates dissatisfied than a fool satisfied."), and advanced by G.E. Moore [146] ("All moral laws are merely statements that certain kinds of actions will have good effects."). It is very effective when we want to quantify the economic values of a decision regarding costs and benefits, such as in healthcare or medical ethics applications, a priority of government funding or wealth redistribution, finance, marketing, risk management, and logistics. However, the theory does not absolutely respect the intrinsic value of anything, such as human life, a person's esteem, and human dignity. Anything can be put on a scale to be weighted for its utility values. If Betham's utility theory is to argue between the good and bad of our decisions, Kantian ethics or deontology is considered to be about right and wrong.

---

[6] We use ethics and morals interchangeably. They are codes of conduct in the workplace and professionals. However, the subtle differences are that moral refers to a person's the best intentions while ethics mean "way of living" in a particular society, which consists of a set of specific rules and behaviours.

*5.1.2. Deontology Theory (Moral and Duty)*

Kant's central notion of ethics is known as the "categorical imperative" [147], which implies the ultimate commandment of reason unambiguously from one's duties and obligations. The Kantian golden ethics rule is "Do unto others as you would have them do unto you." It emphasizes that when we, as the free-human agent, need to make our choices, we should make them responsibly by asking ourselves: "What do we like to be treated in the same way as we treat someone else?" What we treat other people is how we expect to be treated.

This version of ethics is much more powerful than Bentham's utilitarianism because it argues that certain things are so critical, such as human life or dignity, that we must take care of them. On the other hand, certain things are such inferiors, which we must never do because we never want to be done to us. Unfortunately, the categorical imperative does not give us a clear standard or rule on how to decide which is the right description. For example, someone can build a drone to stop criminals, but the same drone might also be thought of as harming a human being. It is identified as "an agent ought to do A and ought to do B and cannot do both". [148] Although there is a counterargument known as "ought implies can" (OIC) in ethics, we could investigate this ethical dilemma from a natural law perspective.

*5.1.3. Natural Law Theory of Ethics*

This type of ethical theory is to study how to profoundly enrich human nature by relying upon anthropology. It is about human characteristics. This ethical theory studies how to fulfil the person, bring the person to be excellent, and nourish human life. It is ethics associated with nature and natural law tradition. This tradition has its root in ancient Greek philosophy found by Socrates, Plato, and Aristotle. Socrates approached ethics by questioning people about what virtues are. Plato followed Socrates' path through his Dialogues. Aristotle pursued the same goal by looking at human nature through an anthropological lens. In contrast to Bentham's utilitarianism and Kantian deontology (duty study), Aristotle offered ethical principles originating in human nature and the virtues (moral excellence) that constitute excellence in human life. To this extent, the natural-law theory of ethics is very close to the virtue-based theory.

Aristotle's Nicomachean Ethics [149] exhibits a moral theory that primarily depends on eudaimonia (happiness), a combination of well-being, happiness and flourishing. Aristotle argued that the general strategy for human life is to pursue what they think will make them happy. However, people often disagree about what is formulated as happiness. Aristotelian ethics seek common patterns of moral excellence regardless of cultural backgrounds and social history. Aristotle's ethics studies how to make humans being prosperous. Thus, it is the natural law theory of ethics. Lobban [150] argued that ethics is anthropology when academic researchers seek approval for their research proposals from institute review boards. The critical problem of anthropological ethics is that it cannot solve some immediate issues. Perhaps, the most effective approach to deal with the issues of TAI ethics is to combine all three ethical theories when new AI technologies challenge some of our fundamental notions of ethics for strategic decision-making.

Practically, when we intend to apply three ethical theories for TAI, we often find many ethical terms (e.g., right, wrong, good, bad, fair, unfair, just, unjust, and trustworthy) and AI concepts (e.g., smart, advanced, and intelligent do not provide specific and measurable definitions) are vague and evolving. Consequently, these terms cultivate different paradoxes. The paradox arises because 1.) a set of inconsistent factual statements, 2.) seemingly reasonable assumptions with acceptable reasoning but an obviously false conclusion, and 3.) an unacceptable conclusion derived from seemingly good premises and reasoning.[151]

*5.2. Paradoxes of the Ethics AI*

Powers and Ganascia [152] asked a similar question: how to approach ethical AI, which they call "the ethics of the ethics of AI," or meta-ethics that studies the nature, foundations and scope of ethical principles and values for the ethics of AI. They examine this topic from five perspectives: 1.) conceptual ambiguities, 2.) the estimation of risks, 3.) implementing machine ethics, 4.) epistemic issue of scientific explanation and prediction, and 5.) oppositional versus systemic ethics approaches. All these challenges have not had practical resolutions yet because "when we turn to the epistemology of the ethics of AI, we find that the ethics of AI will depend on the very science that AI produce." Logically, the meta-ethics idea is a self-reference. It is similar to a barber paradox derived from Russell's paradox.[153]

Using "conceptual ambiguities" as an example, suppose we want to address the ethics of AI. We want to understand and clarify critical terms first, such as agency, responsibility, intention, autonomy, fairness, moral status, and interest, because AI challenges these ethical concepts. However, the concept of AI itself is under rapid development. For example, rule-based systems, chatbots, and expert systems represented AI from the 1970s to the 2000s. Now, the AI agent becomes facial recognition and machine learning. This phenomenon is also known as the **AI effect paradox or "Turing Test" threshold**. The fundamental issue is "What AI is the real AI and what is not?" The answer is "not so straightforward." [154] When we discuss the meta-ethics of AI, we mean principles of "moral principles for AI." The issue is that AI is an emerging technology. It is always ahead of ethical principles. The question is how to decide the moral principles of AI that emerge in the first place by later the moral principles. It is an **AI Ethical paradox.** In short, the real challenge of ethics AI is how to solve a meta-paradox problem. (Refer to Figure 12)

Addressing these paradoxes may require a multi-stakeholder approach that involves AI systems' developers, users, government regulators, and social organizations. This approach should be grounded in ethical principles and guided by ongoing dialogue and collaboration between these stakeholders. It also requires ongoing monitoring and evaluation of AI systems to ensure that they behave in ways consistent with the meta-decisions or our principles.

In addition to the meta-paradox of AI Effect and Ethics, we also face the decision-making paradox [155], also known as the rational paradox. We intend to use the multi-criteria and rational analysis of available information to make the best decision. However, our decision is often influenced by emotions, beliefs, perceptions, and even biases. It results in a self-defeating astray

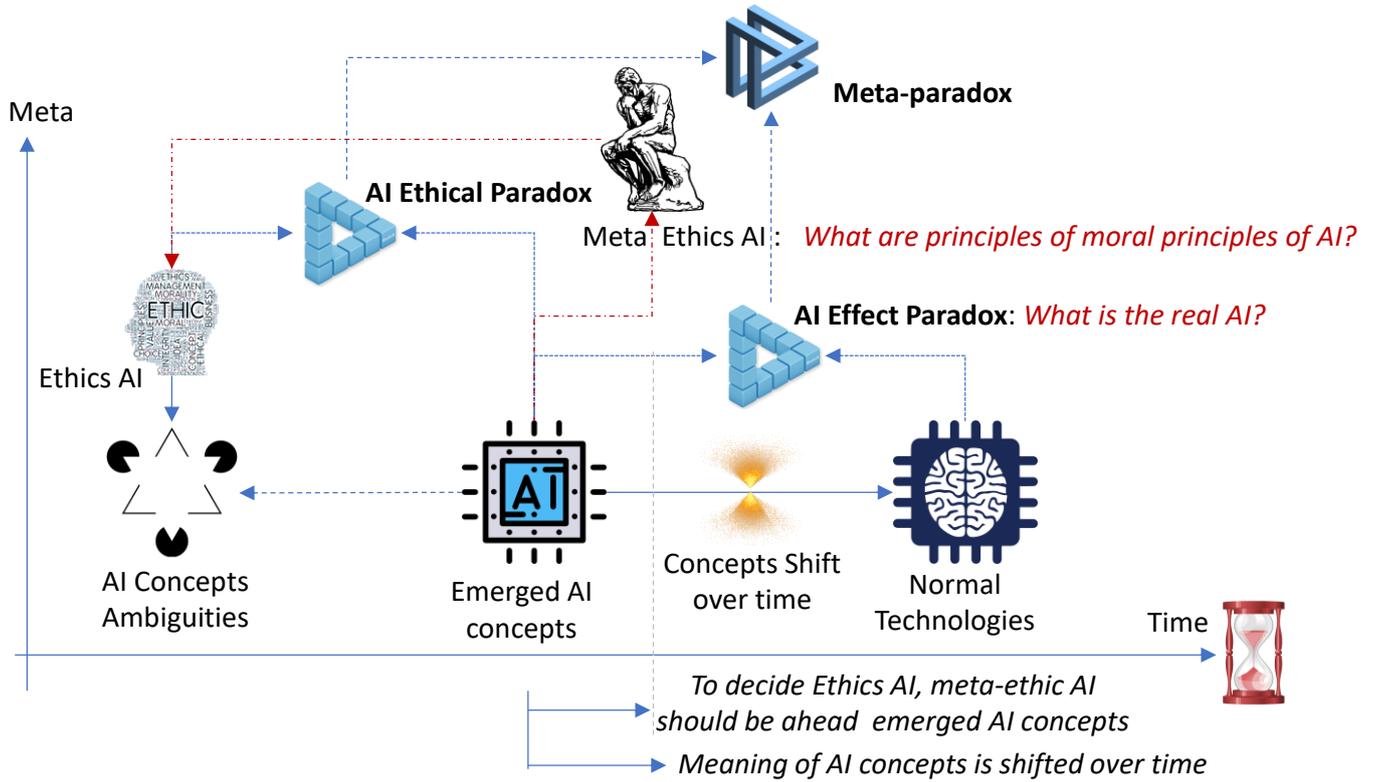

Figure 12: Meta-paradox of Ethics AI for TAI

## 5.3. Machine Learning Paradoxes

In addition to the **AI effect paradox**, one of the well-known ML paradoxes is **Simpson's paradox** (or mix effects), which often happens at different levels of data aggregation. The phenomenon of the Simpson paradox shows that the direction of a statistical result is inverse in the whole dataset compared to sub-datasets. Identifying whether Simpson's paradox imbeds in a dataset for decision-making is crucial, especially for federated learning. Shmueli and Yahav [159] introduce the structure of classification and regression trees to detect potential Simpson paradoxes in a dataset automatically. However, Armstrong and Wattenberg [160], "there is no silver bullet to account for the Simpson paradox." We can only be aware of this paradox when we apply FL for privacy.

Besides Simpson's paradox, there is **Braess's paradox,** which means that if we add more resources to a system, we expect to see increasing AI systems' efficiency. Counterintuitively, the Braess paradox shows the opposite result for some systems, such as adding alternative routes in a traffic network system. Tumer and Wolpert [161] provided the solution of collective intelligence for autonomous driving cars. This paradox often occurs when we adopt reinforcement learning.

When we train a prediction model via a dataset, our intuitive goal is that we want the model to be as accurate as possible. However, accuracy is not the only good indicator for pattern recognition, information retrieval, object detection, and classification problems because we must balance precision [7] and recall (sensitivity) [162] in a training dataset. If a training model gives us nearly 100% of prediction accuracy, it could mean the selection of the training dataset is imbalanced. It is harmful.[163] The near-perfect accuracy could be misleading because the correct metric for prediction power is equal to the precision. For a medical diagnosis, both sensitivity [8] and specificity [9] are essential. In other words, **the accuracy paradox** exhibits that perfect accuracy is not always desirable. A good training model needs to include false instances to train itself to differentiate between true and false.

The learnability of a learning model is undetermined [164] because Godel's incomplete theorems showed that not everything is provable by logic alone. Machine Learning has the same fate because the foundation of ML is logic. The study [164] demonstrated that "estimating the maximum" (EXM) learnability cannot be proved or disproved. That is to say that AI/ML could not solve some problems. The paradox is that ML is unlearnable, known as **the learnability paradox**.

The last well-known paradox is the so-called **Moravec's paradox** [156], originated by Hans Moravec, Rodney Brooks, and Marvin Minsky. It states that machines can easily accomplish tasks (complex logic reasoning) that humans find very difficult to do. Conversely, some common sense activities (such as tieing up shoelaces) that are effortless for human is highly challenging for machines, such as strategic intuition or gut feeling, creativity and imagination, moral reasoning, emotional intelligence, consciousness, common sense, and practical wisdom. Even if a machine can do some human activities, it takes much computational power. Perhaps, common sense ethics is complex for machines to grasp. Overall, we highlight eight ML paradoxes.

---

[7] positive predictive value or PPV = true positives/(true positives + true negatives)
[8] Recall = Sensitivity = true positives /( true positives + false negatives)
[9] Specificity = true Negatives/(true negatives + false positives)

*5.4. TAI Paradoxes*

Likewise, TAI can also involve several paradoxes that must be addressed to build trust and confidence for many AI systems' users. We have already mentioned some dilemmas in the above survey (See Figure 9). Here, we want to highlight some typical examples of TAI paradoxes: 1.) **The Paradox of Autonomy** is the relationship between fully autonomous AI systems and machines' understanding of reality. Currently, machines are very challenging to understand and control in reality, especially if we want AI systems to behave ethically, empathically, consciously, and fairly as humans do. 2.) **The Transparency Paradox** is another TAI challenge because the more transparent AI systems are, the more likely they are to be susceptible to hacking or misuse. The issue is how we can make a balance or trade-off between security and transparency or an explanation. 3.) **The Bias Paradox** is often the primary issue for TAI. We hope that ML algorithms or AI systems can be trained by unbiased data, but collected datasets are often biased in ways that reflect existing social prejudice and cultural influence, which can perpetuate discrimination and inequality. 4.) **The Paradox of Human Control** implies that while humans desire to control AI systems, too much control or regulations could stifle AI innovation and prevent AI systems from reaching their full potential. 5.) **AI explanation (XAI) paradox** is perhaps the most challenging issue for TAI. It is closely related to the transparency problem. However, The paradox of explainability also refers to the fact that when AI systems become more complex, their decision-making processes become less transparent and more difficult to explain. 6.) **The Paradox of Accountability** is similar to the AI Ethic paradox, in which TAI should be accountable for its decisions and actions, but the lack of clear legal frameworks and standards for AI makes it difficult to establish responsibility. 7.) **The Paradox of Scale** shows that when we build AI systems, we want the systems to be scalable to process ever-increasing large data volume. However, the larger the scale of the AI system, the more difficult it becomes to ensure that it operates in a trustworthy manner.

*5.5. Paradoxes of Strategic Decision-Making*

This study mainly focuses on TAI but is oriented toward strategic decision-making. Consequently, we want to highlight some paradoxes of strategic decision-making that could intertwine with both ML and TAI paradoxes. There are at least **20 different paradoxes** when we make strategic decisions. Here, we only list seven paradoxes closely associated with TAI and ML. **1.) The exploration-Exploitation paradox** can also be interpreted as the K-armed or multi-armed bandit problem[157] that is closely tied to reinforcement learning. It answers the question of how to deploy ML algorithms. From a strategic decision-making perspective, this paradox involves the tension between exploring new options and exploiting known ones. The paradox arises because investing too much in exploration may lead to high costs and low returns, while investing too much in exploitation may lead to missed opportunities. **2.) The Short-Term and Long-Term Paradox** concerns reconciling between short-term (or tactic) gains and long-term (or strategic) goals (objectives). As we should see, focusing on short-term gains can provide immediate benefits, but focusing on long-term goals can lead to sustained success. The paradox arises because emphasizing too much on tactic gains may result in sacrificing strategic objectives, while concentrating too much on long-term goals may lead to abandoning short-term gains. This paradox is closely related to the dimension of sustainability of TAI under the long-term prosperity model. **3.) The Strategy and Execution Paradox** involves the stress between formulating a strategy and executing it effectively. It is also related to the sustainability dimension of TAI under resource management models. From a strategic planning perspective, formulating a strategy can provide the right direction and focus. However, devoting too much effort to planning can lead to neglecting the strategy of execution. Conversely, dedicating too much energy to strategic planning could fail to commit enough resources to strategy execution **4.) The Rationality and Intuition Paradox** indicates tension arises due to whether using rational analysis and logic or intuition and emotional intelligence to make strategic decisions. A rational approach can lead to logical and data-driven decision-making, but employing intuition can lead to creativity, innovation, and empathy. The paradox is triggered by too much rationality may result in a lack of creativity and imagination, while too much reliance on intuition may create unrealistic fantasy. This paradox is related to the XAI of the reason-based implementation model. **5.) The Stability and Flexibility Paradox** express a clash between maintaining stability and adapting to change. Maintaining stability can ensure predictability and security while adapting to change can provide growth and competitive advantage opportunities. The paradox emerges because too much stability may cause stagnation and missed opportunities, while too much flexibility may produce instability and a lack of direction. The paradox is highly related to the TAI of the robustness of dimension. **6.) The Competition and Collaboration Paradox** addresses the conflict between competition and collaboration. We know that competition will generate innovation and differentiation. However, collaboration will yield the benefits of knowledge-sharing and joint problem-solving. The paradox occurs that too much competition could create a hostile environment, while too much collaboration may result in complacency and lack of creativity. This paradox is related to the sustainability dimension of TAI. **7.) The Openness and Secrecy Paradox** refers to the competing interest between being open and transparent or being secretive and protecting confidential information. Being open and transparent can build trust and foster collaboration. However, being secretive can protect intellectual property and sensitive information. The paradox stirs up if being too open may reveal confidential information and vulnerability, while being too secretive may lead to a lack of transparency and trust. This paradox is across the safety and transparency dimensions of TAI.

## 6. Summary, Conclusion and Future Direction

The research question is, what has been done regarding TAI in the last two decades? We need to understand how we can trust the result of machines or how to make meta-decisions, especially for strategic decision-making, because we often face an overwhelming amount of information, and many pieces of evidence are contradicted or paradoxical.

Researchers have studied the topic from different perspectives for years. It allows us to collect over a thousand papers, books, dissertations, AI proposals, and industrial reports. In order to review the enormous amount of literature within a limited timeframe, we formulate the TAI framework that consists of three domains and ten dimensions, which aligns with Marr's computational theory of vision. We then generate the selection criteria for the collected literature. 1.)The topic of selected research papers must be within the scope of three domains and ten dimensions, 2.)The study should include specific implementable algorithms and processing principles.

3.)The selected literature should provide empirical results and highlight limitations to compare them with the current state of the art. 4.) The study must also present a research method that can be validated.

*6.1. Objectives Level*

We developed the novel TAI taxonomy or framework with three domains and ten dimensions to achieve the computational objectives for this survey. The objectives are the strategic goal of a decision-maker. A specific business goal may include revenue growth, market penetration rate, cost of production, risk management, and long-term sustainability. To compute these objectives, such as 10% revenue growth with a 7% cost reduction in the next five years, we identify different domains (trust levels) to understand how much we can trust the result produced by ML because revenue growth and cost reduction often contradict each other. Sometimes, the strategic objectives are paradoxical. This issue leads to deciding what to decide or a meta-decision of strategic decisions or TAI.

*6.2. Domains Level*

Based on the newly created taxonomy or framework of TAI, we defined three domains of trust: articulate, authentic, and basic (See Figures 2 and 5). These domains reflect domain knowledge and different levels of trust. Articulate trust is trust that has been clearly defined without any ambiguities, which is beyond simple trust (or the "noncognitive" nature of trust [158]). This level of TAI knows the risks and vulnerabilities well when we see the outputs produced by AI/ML. The next level of trust is Authentic trust. It does not deny distrust but accepts, transcends, absorbs, and overcomes it. A typical example of this level of trust is the security or safety AI, such as hacking, ransomware, password attacks, and malware. Basic trust is a baseline belief without saying. We often take it for granted. It is a natural level of trust. From the TAI perspective, basic trust is like setting basic or default parameters for a strategic objective. One of the typical dimensions to underpin the objectives of TAI is that we assume the results produced by AI systems or ML can be reproducible, reliable, and robust. Although each level of trust may correspond to three different dimensional classes, each class is just a placeholder. We can select different dimensions to form different levels of trust. It is dependent on TAI objectives.

*6.3. Dimensions, Implementation Models and Schemes Level*

According to Figures 2 and 5, we create ten dimensions, 33 implementation models and 106 schemes/tools (60 plus 46 schemes, See Figures 5 and 8) to deliver any specified TAI outcomes. If the ethics theories are driven from top to bottom, then the implementation models, schemes/tools are driven from bottom to top. It is data-oriented (Refer to Figure 2). We can deploy different models for the same goal. Consequently, we could encounter many dilemmas, such as "interpretation and explanation", "persuasion and fact-based transparency", and "manipulation and evidence-based explanation" (See Figure 9). Moreover, AI/ML is dynamically evolving over time. There are many levels of paradoxes regarding AI, Ethics AI (As shown in Figure 12), and strategic decision-making. The notions of trust and trustworthiness are also a dilemma because there will be no trustworthiness without initial trust. TAI is a multidisciplinary topic associated with computer science, cognition science, economics, psychology, social science, and philosophy. (See Figure 11). This survey primarily focuses on computer science by shedding light on the philosophical morals and ethics of AI. We intend to provide a comprehensive survey of TAI for a meta-decision problem.

*6.4. Future Direction*

In future research, we will consolidate this TAI framework from a strategic decision perspective and implement the various TAI algorithms with defined domain vectors on a cloud or HPC platform. The critical research problem is transforming some ML, ethics AI, and strategic decision-making paradoxes or dilemmas into solvable problems for desired TAI objectives.

ACKNOWLEDGMENTS

This research was funded by the Luxembourg National Research Fund (FNR), grant reference C21/IS/16221483/CBD. For open access, the author has applied a Creative Commons Attribution 4.0 International (CC BY 4.0) license to any Author Accepted Manuscript version arising from this submission."

# Appendix

Table 1: Summary of data governance models[117]

| Model | Applications | Goals | Value | Principles |
|---|---|---|---|---|
| Data Sharing Pools (DPs) | Business entities Public bodies | Fill knowledge gaps through data sharing  Innovate and develop new services | Private Profit Economic Growth | Principle of 'data as a commodity Partnerships Contracts (e.g. repeatable frameworks) |
| Data cooperatives (DCs) | Civic organizations Data subjects | Rebalance power unbalances of the current data economy Address societal challenges Foster social justice and fairer conditions for value production | Public interest Scientific research Empowered data subjects | Principles from the cooperative movement Data commons 'Bottom-up' data trusts GDPR Right to data portability |
| Public data trusts (PDTs) | Public bodies | Inform policy-making Address societal challenges Innovate Adopt a responsible approach to data | Public interest More efficient public service delivery | Principle of 'data as a public infrastructure' Trust building initiatives Trusted intermediaries Enabling legal framework |
| Personal data sovereignty (PDS) | Business entities • Data subjects | Data subjects self-determination Rebalance power unbalances of the current data economy Develop new digital services centred on users' need | Empowered data subjects Economic growth Private profit Knowledge | Principle of 'technological sovereignty.' Communities and movements (e.g. MyData) Intermediary digital services (personal data spaces) GDPR Right to data portability |

Table 2: Comparison of TAI Schemes/Tools

| Dimension type | Authors | Ref | Year | Main Contributions | Methods | To be Improved |
|---|---|---|---|---|---|---|
| Explainability and Transparency | Sharma et al. | [57] | 2019 | a new unified model: CERTIFAI, for a counterfactual explanation. The first time proposed method to how to measure a black-box model | counterfactual method and customized GA | Speed up the GA process |
| | Wachter et al. | [58] | 2017 | a lightweight form of unconditional counterfactual explanations | Neural Net, SVM, Regression | How to quantify a sufficient and relevant set of counterfactuals |
| | Byrne | [59] | 2019 | Demonstrate that not all counterfactuals are helpful | neural network | XAI to combine with psychological experiments |
| | Rudlin | [70] | 2019 | Identify reasons why XAI for strategic decision is a bad practice | Accurate, interpretable models | Constructing optimal logical models, optimal sparse scoring systems, defining interpretability |
| | Burns et al. | [73] | 2020 | Reframe the black box as a multiple-hypothesis testing problem to discover important features | Interpretability Randomization Test and One-Shot Feature Test | how to choose which subsets of features to test |
| | Gilpin | [74] | 2019 | Identify insufficiencies of current approaches for XAI | Exploring approach | How to adopt a systematic approach |
| | Herman | [77] | 2017 | Proposed descriptive and persuasive explanations. | Exploring approach | how to avoid just relying on a narrative approach |
| | Bills et al. | [78] | 2023 | AI to explain AI | Three steps model | assumption of the black box can explain black box need further consolidation, using Non-negative matrix factorization (NMF) and Singular Value Decomposition (SVD) or dictionary learning approaches to work around polysemantic |
| | | | | | | |
| **Fairness and Diversity** | Mahoney et al. | [79] | 2020 | AI Fairness 360 toolkit how to measure bias and remove it | 21 mathematical definitions of fairness | how to tackle fairness from multiple stakeholders and perspectives |
| | Madaio et al. | [80] | 2022 | Identify impacts on fairness work due to lack of domain experts, engagement with direct stakeholders | Inquiry methods | Limited sample size and representation |
| | Zhou et al. | [81] | 2021 | information-lossless de-biasing technique | generate synthetic data | how to avoid prediction accuracy sacrifice |
| | Grgic-Hlaca et al. | [82] | 2018 | Proposed a procedure fairness measurement | Three new scalar measurements of fairness | The experiment results do not hold in general |
| | Mozannar et al. | [83] | 2020 | learning fair predictions in a setting where protected attributes are only available | Two steps learning | The solution can not learn predictors in the absence of any demographic data |
| | Bellamy et al. | 83 | 2019 | Introduction of open source Python toolkit for algorithm fairness | The architecture of the package | how to extend the variety of types of explanations |
| | | | | | | |
| **Generalizability** | McDermott | [87] | 2019 | Provide metrics for reproducibility in ML for humans regarding three lenses technical, statistical and conceptual | Analytic approach | how to test the framework |
| | Azad | [88] | 2021 | Highlight common errors that lead to poor reproducibility and generalizability | leveraging proposed taxonomy | how to implement and measure techniques to increase generalizability |
| | Arjovsky | [89] | 2019 | Propose invariant Risk Minimization (IRM) | thresholding statistical hypothesis tests | limited application, high computational cost, difficulty in selecting the appropriate penalty functions, sensitivity to the choice of hyperparameters, difficulty in interpreting the learned models |
| | Ahuja | [90] | 2020 | Propose invariant Risk Minimization (IRM) based on game theory | game theory | how to apply the proposed solution for invariance in causal inference |
| | Kamath | [91] | 2021 | Demonstrate that IRM can be brittle | Analytic method | what is an alternative solution? |
| | | | | | | |
| **Privacy** | Yu | [95] | 2006 | Privacy-Preserving SVM classification | Global SVM classification | data must be horizontally partitioned |
| | Karr | [96] | 2009 | A protocol can estimate coefficients and standard errors of linear regression for vertically partitioned data | | how to share nonlinear analysis securely |
| | Li | [98] | 2020 | A novel boosting framework to allocate the privacy budget (Differential Privacy Boost or DPBoost) | Gradient Boosting Decision Tree (GBDT) and light gradient boosting machine | issues of interpretability, scalability, and overfitting |
| | Sweeney | [99] | 2002 | K-anonymity | Preferred Minimal Generalization Algorithm | risk of re-identification attack, computational complexity, and difficulty in handling data updates |
| | Aggarwal | [100] | 2005 | The effect of dimensionality on k-anonymity for high dimensionality cases | k-anonymity | Difficulty in achieving high levels of anonymity and handling sparsity, computational complexity, and limited effectiveness against re-identification attacks |
| | Kasiviswanathan et al. | [104] | 2016 | Efficient Private Empirical Risk Minimization for high-dimensional learning | empirical risk minimization, perturbation techniques | overfitting, difficulty in feature selection, sensitivity to hyperparameters |
| | Andrew et al. | [105] | 2021 | Differentially Private (DP) learning with adaptive clipping | DP federated averaging with Federated learning | increasing computational complexity, difficulty in achieving good accuracy, difficult in selecting the appropriate clipping norm, limited protection against membership inference attacks |
| | Chaudhuri et al. | [106] | 2011 | Differentially Private (DP) Empirical Risk Minimization | Empirical Risk minimization | increasing computational complexity, difficulty in achieving good accuracy, sensitivity to hyperparameters, difficulty in scaling to large datasets |
| | Abadi et al. | [107] | 2016 | Differential private stochastic gradient descent (PD-SGD) | Stochastic Gradient Descent | increasing computational complexity, difficulty in achieving good accuracy, sensitivity to hyperparameters, difficulty in scaling to large datasets |
| | Xu et al. | [108] | 2019 | VerifyNet for secure and verifiable Federated learning | Double masking protocol | Limited support for non-iid data, increasing communication overhead of the entire protocol |
| | Truex et al. | [109] | 2019 | A hybrid approach of Differential Privacy (DP) and Secure Multiparty Computation | DP and SMC | high computational cost, sensitivity to the choice of privacy parameters, difficulty in tuning |
| | | | | (SMC) in Federated Learning | | hyperparameters |
| | Teng & Du | [110] | 2019 | hybrid multi-group | random and secure multi-party computation, ID3 Decision Tree | Fixed window size for decision tree building, limited ability to capture complex relationships, sensitivity to the choice of window size, limited adaptability |
| | Tramer et al. | [111] | 2017 | Identify the vulnerability of adversarial training and proposed ensemble adversarial training scheme | adversarial training | elaborate black box attacks |
| | | | | | | |
| **Data Governance** | Eryurek | [112] | 2021 | A practical guide on how to implement and scale data governance incorporates the ways people, processes and technology work together to ensure data is trustworthy | monitoring, agility, and planning | of some content need to be further consolidated. Different data governance approaches should be presented, not only contributed by Google |
| | Gritsenko | [113] | 2022 | Propose a design-based governance mode | algorithms in governance | bias issue because algorithms are only as good as the data they are trained on, lack of transparency, limited scope, ethical issues |
| | Kooiman | [114] | 2009 | Present meta-government as an integral part of governance | Analytic | complexity, limited impact, and accountability |
| | de Bruijn and Janssen | [116] | 2017 | Highlight a few challenges in cybersecurity and offer high-level strategies | Narrative discussion | practical solution |
| | Micheli et al. | [117] | 2020 | Highlight four models of data governance: data sharing pool, data cooperatives, public data trusts, and personal data sovereignty | Analytic | practical solution |
| | | | | | | |
| **Safety** | Amodei et al. | [118] | 2016 | Highlight five accident safety issues due to poor design of real-world AI systems | analytic in reinforcement learning | how to gauge the risk of larger accidents |
| | Leike et al. | [119] | 2017 | Present a set of reinforcement learning environments for various safety properties | study cases | interpretability and formal verification |
| | | | | | | |
| **Reliability & Robustness** | Saria | [126] | 2019 | Provide an overview of machine learning reliability issue and draw connections to other trust AI concepts and highlight novel approaches for measuring reliability | Analytic | How to solve ML reliability issues in practice |
| | Brendel | [127] | 2018 | Highlight the most neglected sub-category of adversarial attacks: a decision-based attack that triggers ML reliability & robustness issue for deployment | boundary Attack with a random walk | on improving memory for its previous steps, fixed size steps. The human decision is not purely random |
| | Moustapha et al. | [128] | 2022 | Propose a generalized modular framework to build on-the-fly efficient active learning strategies for 20 reliability benchmark problems | active learning | How to understand the effect of the threshold in the stopping criterion, more advanced learning function: reinforcement, semi-supervised, transfer, unsupervised, deep, and Bayesian learning |
| | | | | | | |
| **Sustainability** | Al-Surmi | [129] | 2022 | Propose a novel three-phase decision-making framework via AI processes for operational efficiency | Survey | More details of AI for the decision-making process |
| | Vinuesa et al. | [130] | 2020 | Provide a high-level summary of sustainable development goals (environment, economy, and Society) and 169 targets influenced by AI | the consensus-based expert elicitation process | the presented analysis represents the perspective of the authors |
| | Lacoste et al. | [133] | 2019 | Machine Learning Emissions Calculator | Web-based calculation | transparency of calculation |